\documentclass[lettersize,journal]{IEEEtran}
\usepackage{amsmath,amsfonts}
\usepackage{lineno,hyperref}
\usepackage{array}
\usepackage[caption=false,font=normalsize,labelfont=sf,textfont=sf]{subfig}
\usepackage{textcomp}
\usepackage{stfloats}
\usepackage{url}
\usepackage{verbatim}
\usepackage{graphicx}
\usepackage{algorithm,algpseudocode}
\usepackage{algorithmicx}
\usepackage{glossaries} 
\usepackage{xcolor}
\usepackage{cite}
\usepackage{amssymb}
\usepackage{latexsym}
\usepackage{amsmath,amsfonts,amssymb,amsthm}
\usepackage{float}
\usepackage{cite}
\usepackage{array}
\usepackage{booktabs}
\usepackage{multicol}
\usepackage{multirow}
\usepackage{mathptmx} 
\usepackage{relsize}
\usepackage{fancyhdr}
\usepackage{times} 

\begin{document}

\title{\emph{Advanced Acceptance Score}: {A} {H}olistic {M}easure for {E}valuating {G}esture {B}iometric {Q}uantification}

\author{Aman Verma, \thanks{Aman Verma is with Dept. of Electrical Engineering, IIT Delhi, India.}Seshan Srirangarajan, \thanks{Seshan Srirangarajan is with the Dept. of Electrical Engineering and Bharti School of Telecommunications at IIT Delhi, India}, \IEEEmembership{Senior Member, IEEE}, and Sumantra {Dutta~Roy}, \thanks{Sumantra {Dutta~Roy} is with Dept. of Electrical Engineering, IIT Delhi, India.}\IEEEmembership{Senior Member, IEEE}  }
\maketitle
\begin{abstract}
Quantifying biometric characteristics within hand gestures involve derivation of fitness scores from a gesture and identity aware feature space. 
However, evaluating the quality of these scores remains an open question. 
Existing biometric capacity estimation literature relies upon error rates. 
But these rates do not indicate goodness of scores.  
Thus, in this manuscript we present an exhaustive set of evaluation measures. 
We firstly identify ranking order and relevance of output scores as the primary basis for evaluation. 
%
%
In particular, we consider both rank deviation as well as rewards for: (i) higher scores of high ranked gestures and (ii) lower scores of low ranked gestures. 
We also compensate for correspondence between trends of output and ground truth scores. 
Finally, we account for disentanglement between identity features of gestures as a discounting factor.
Integrating these elements with adequate weighting, we formulate advanced acceptance score as a holistic evaluation measure. 
To assess effectivity of the proposed we perform in-depth experimentation over three datasets with five state-of-the-art (SOTA) models. 
Results show that the optimal score selected with our measure is more appropriate than existing other measures. 
Also, our proposed measure depicts correlation with existing measures. This further validates its reliability. We have made our \href{https://github.com/AmanVerma2307/MeasureSuite}{code} public.
\end{abstract}
\section{Introduction}
\label{sec:intro}
\begin{figure*}[!htb]
    \centering
    \includegraphics[width=1.0\linewidth]{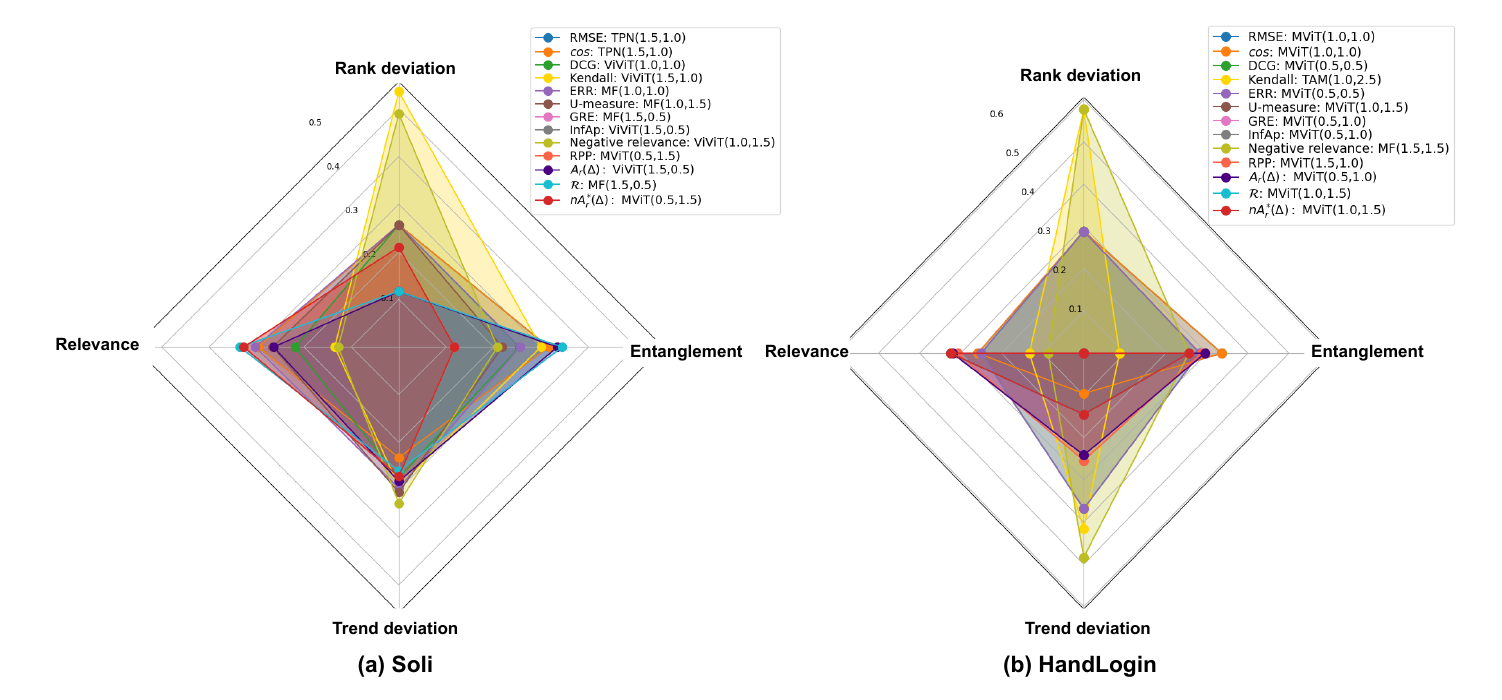}
    \caption{\textbf{Comparison of the proposed $nA_{r}^{*}(\Delta)$with existing ranking and retrieval measures.} 
    We compare the optimal score suggested by the measures in terms of four design criteria: (i) deviation in ranking order of gestures, (ii) entanglement between biometric traits of different gestures, (iii) trend deviation, and (iv) quality or relevance of quantification estimates. 
    For optimal score, the first three are expected to be minimal, while relevance is to be maximized. 
    We have normalized the values in the range between $[0,1]$.    
    This analysis reveals that score values selected by $nA_{r}^{*}(\Delta)$ jointly satisfies all the design criteria.  
    This is not observed for other measures. This merit can be accredited to multi-facet and task specific evaluation.}
    \label{fig:MS_Comp_Radar}
\end{figure*}
Hand gesture biometrics hold applications in personalized human computer interactions~\cite{mendels2014user, kong2020continuous}.
Since it offers ease of collection and multimodal authentication, several works have recently emphasized on hand gesture as biometrics~\cite{song2023understanding, zhang2024robust, xie2023msba}. 
However, limited attention has been paid to find out the gestures that are better for the purpose. 
To fulfill this gap, \cite{verma2024quantifying} introduced the DGBQA framework. 
In method, DGBQA outputs scores corresponding to each gesture, that quantifies biometric goodness. 
For next step towards practical deployment, the DGBQA framework must pass quality standards. To this end, it is essential to rigorously evaluate framework's performance. Hence, in this manuscript, we propose task-specific and holistic evaluation measures for hand gesture biometric characteristics quantification.

The preliminary choice for evaluating any biometric system are the matching rates. However, they are not designed to assess score values.  
Also, the existing biometric capacity estimation literature \cite{terhorst2022limited, boddeti2023biometric} had not proposed any particular evaluation measure. 
\cite{verma2024quantifying} modeled the evaluation as a ranking problem. 
To be specific, authors reported results in terms of rank deviation: the difference in the ranking orders of output and ground truth scores. 
They also proposed the ICGD score, which measures the amount of disentanglement left between identity features of different gestures. This entanglement is introduced because of joint gesture and identity aware characteristics of the feature space. 
The more ICGD score is, the less reliable are the output scores. 
Although with these two measures, authors were able to differentiate between high and low quality scores. We find that optimizing them is necessary condition, but not sufficient. 
There are other important facets of evaluation. 
For instance, a high quality score values must ensure that there should be higher scores for high ranking gestures, and lower scores for low ranking gestures. Also, the difference of output scores between two consecutively ranked gestures must be proportional to that observed in the ground truth scores. 
We also observed that lower ICGD score did not guarantee lower rank deviation (vice-versa is also true). Thus, either of these does not completely represent the fitness of the scores. 
This raises the requirement to club them together with other facets, and obtain a holistic evaluation measure. To this end, we identify the essential criteria for evaluation:

\begin{enumerate}
    \item \textbf{Rank deviation:} This accounts for the amount of difference in ranking order as per the DGBQA and ground truth scores.
    
    \item \textbf{Relevance:} For high relevance, the DGBQA score values should be high for high ranking gestures. While, for low ranking gestures, they should low.
    
    \item \textbf{Trend deviation:} This represents the difference in score value progression of DGBQA and ground truth scores. Our hypothesis is that there should be proportionality in the difference between DGBQA and ground truth scores of two consecutively ranked gestures. Lesser the proportionality, more the value of trend deviation.

    \item \textbf{Entanglement:} This is a property of feature space. It measures the amount of entanglement remaining between identity features of different gestures. It can be estimated using the ICGD score.
\end{enumerate}

\vspace{0.20cm}

\noindent Verma et al. \cite{verma2024quantifying} have already proposed measures for rank deviation and entanglement. 
Thus, the research gap lies in formulating measures for the remaining two criteria. 
Evaluation measures for assessing the grade of retrieved results in information retrieval has been substantially studied \cite{ochoa2008relevance, chapelle2009expected, sakai2013summaries, lioma2017evaluation, gienapp2020estimating, maistro2021principled, li2024evaluation}. 
This is proximal to our relevance criterion. Discounted cumulative gain (DCG) \cite{jarvelin2002cumulated} from the same domain has now become the industry standards. 
More recently, several other flavors of DCG, such as the one including negative grading \cite{gienapp2020impact}, findability \cite{sinha2023findability}, topic difficulty \cite{gienapp2020estimating}, and credibility \cite{lioma2017evaluation} have been proposed. 
However, these measures reward only high grades of high ranking scores, while discount the low grades of low ranking scores. This is contradictory to our desiderata. 
Apart from DCG, several other works have considered rank deviation \cite{lioma2017evaluation, valcarce2020assessing} or both rank deviation and grading \cite{chapelle2009expected, moffat2008rank, amigo2022ranking}.
These works however do not parametrize grading with the rank. 
This limits them to reward a wider score value spectrum in comparison to a squeezed one. To cater the above discussed concerns, we propose relevance measure ($\mathcal{R}$). In particular, it considers ranking order to determine whether to reward high scores of high ranking gesture, or low scores of low ranking gestures. 
Alongside relevance, we also design trend match distance ($\Psi$) to capture trend deviation. 
It utilizes the difference in DGBQA score values between two consecutively ranked gestures to estimate trends in ground truth score values. 
The estimated values are then compared with ground truth values. 
We further combine the formulated measures into an unified figure of merit. 
This is achieved by considering the relative importance between the design criterion, and then fusing the individual measures with suitable weights. 
We refer the resultant as the advanced acceptance score ($A_r^{*}(.)$), or its normalized version as the normalized advanced acceptance score ($nA_r^{*}(.)$). 
As illustrated in Fig. \ref{fig:MS_Comp_Radar}, the advanced acceptance score selects that score value, which optimally satisfies all the design criteria. 
This highlights its capabilities to perform holistic evaluations. 
Following summarizes our key contributions:

\begin{itemize}
    \item We present theoretical basis for evaluating hand gesture biometric quantification. Based on our theory, we highlight four key design criteria: (i) rank deviation, (ii) entanglement, (iii) relevance, and (iv) trend deviation. 

    \item For each of the design criterion, we formulate an evaluation measure. Unlike other score quality measures, our relevance measure rewards both high scores of high ranking gestures, and low score of low ranking gesture.
    
    \item Another novel contribution is the trend match distance for calculating the difference in local trends between the ground truth and output score values.

    \item By allocating adequate weightage to the individual evaluation measures, we aggregate them into an holistic evaluation measure called as advanced acceptance score.

    \item The score values designated optimal by the advanced acceptance score jointly satisfies all the design criterion. We validate this over three datasets on five state-of-the-art architectures. Also, the proposed measure results in more appropriate score selection in comparison to existing retrieval/ranking measures (refer Fig. \ref{fig:MS_Comp_Radar}).
\end{itemize}
The remainder of this manuscript is arranged as follows. 
In the next section, we present a review of existing evaluation measures. After that, in section \ref{sec:proposed} we discuss the proposed advanced acceptance score. 
Finally, in section \ref{sec:experimetal} we present experimental study to prove the effectiveness of the proposed measure. While we conclude the paper in section \ref{sec:conclusion}, highlighting research gaps and scope of improvement.


\section{Related works}
\label{sec:review}

\subsection{Evaluation measures for biometrics and hand gestures}
\label{sec:review_bioGes}
Previous works in hand gesture biometrics \cite{song2023understanding, zhang2024robust, xie2023msba} use matching rates for evaluation. As discussed earlier, these rates are not designed to evaluate score values. Existing capacity estimation literature \cite{terhorst2022limited,boddeti2023biometric,zunino2017revisiting} also utilizes matching rates to qualitatively compare the capacity estimates. 
Their analysis did not involve any figure of merit to assess the effectivity of estimation. 
In contrast, we develop evaluation measures considering error rates as the basis for ground truth. 
Majority of innovation in biometric evaluation measures has been for fairness measurement \cite{de2021fairness, dorsch2024fairness}. In \cite{de2021fairness}, authors proposed to compute fairness as the performance difference between the most and least populated demographic groups. While \cite{dorsch2024fairness} proposed several measures based upon dispersion at inter-intra group level to identify biased demographic groups.
Although these fairness measures are not suitable for our problem, they advocated for task specific and multi aspect evaluation. 
A recent work \cite{van2024novel} evaluated biometric performance in terms of space covered by identities in the feature manifold. Despite being effective, it does not matches the evaluation needs of our problem. 
In gesture analysis literature, gesture recognition models are evaluated in terms of standard accuracy measures \cite{garg2024gestformer, kopuklu2020online}. Since, our feature space is both gesture and identity aware, measuring accuracy of a single task is suboptimal. Moreover, accuracies does not highlight the amount of entanglement in the feature space. 
\cite{yoon2020speech} proposed Frechet Gesture Distance (FGD). It utilizes the difference between Gaussian mean and covariance of ground truth and generated gestures. 
It neither captures entanglement nor the ranking order. 
For judging the alignment between input text and generated motion pattern, \cite{voas2023best} proposed MoBERT score. Again, it is also unsuitable to judge score values.
Thus, this review reveals that the aligned literature has not formulated relevant measures. 

\subsection{Retrieval evaluation measures}
\label{sec:review_rankScore}
Retrieval measures can be categorized into two categories: (i) ranking order based and (ii) score quality and rank based.
The measures that utilize only the rank order \cite{lioma2017evaluation, valcarce2020assessing, yilmaz2008estimating, amigo2022ranking} does not convey the quantity by which the ranks were better or inferior.
One such measure is global ranking error (GRE) \cite{lioma2017evaluation}. It discounts the deviation in lower ranks. This is unsuitable for our problem, as we are interested in computing the total rank deviation in the output scores.
There other ordering based measures such as Levenshtein distance and \cite{amigo2022ranking} capture similar information as rank deviation. 
Specifically, \cite{amigo2022ranking} computes the joint entropy of ranking order and the ranking order derived as per the output scores. However, the rank of DGBQA Scores are themselves derived from the output scores, hence the second part is trivial. 
To overcome shortcomings of rank based measures, the second category of measures integrate score values. The family of discounted cumulative gain (DCG) based measures  \cite{jarvelin2002cumulated, tang2017investigating, gienapp2020impact, gienapp2020estimating, sinha2023findability, jeunen2024normalised} serve as the primary choice.
They compute the total grade in the retrieved results by discounting the grades of the low ranking documents. However, this undesirable for our problem, as we also seek how low were the low ranking scores.
To this end, we use both the score value as well as its inverse, and parametrize them by the rank. Similar to DCG, the other evaluation measures such as expected reciprocal rank \cite{chapelle2009expected}, rank based precision (RBP) \cite{moffat2008rank}, rank based utility \cite{amigo2018axiomatic}, and U-measure \cite{sakai2013summaries} also pay low importance to the low ranking scores. 
Recent evaluation measures such as recall paired preference \cite{diaz2022offline} and diversity intra-list average distance (DILAD) \cite{li2024evaluation} ignores the ranking order for estimating score quality. This again results in discounting of low ranking scores. 

\noindent \textbf{Comparison with our previous work~\cite{verma2024quantifying}}:
The closest work to our proposal is the acceptance score proposed in \cite{verma2024quantifying}.
However, this work advances the former in terms of: (i) \textbf{holistic nature}: unlike acceptance score, the proposed $A_r^{*}(.)$ includes terms for entanglement and trend deviation. 
(ii) \textbf{Stronger theoretical basis}: The previous work~\cite{verma2024quantifying} does not give any theoretical basis for the relevance parameter. In contrast, this manuscript presents a structured theoretical formulation. (iii) \textbf{Rigorous evaluation}: unlike ~\cite{verma2024quantifying}, we also present a full scale evaluation of the proposed over three datasets. In particular, a comparative study with relevant measures (refer Fig. \ref{fig:MS_Comp_Radar}) reveals that our proposed is optimal for satisfying the design criterion. Furthermore, to the best of our knowledge this is the first work to exhaustively validate a measure for gesture biometrics.

\begin{figure*}[!t]
    \centering
    \tiny
    \includegraphics[scale=0.58]{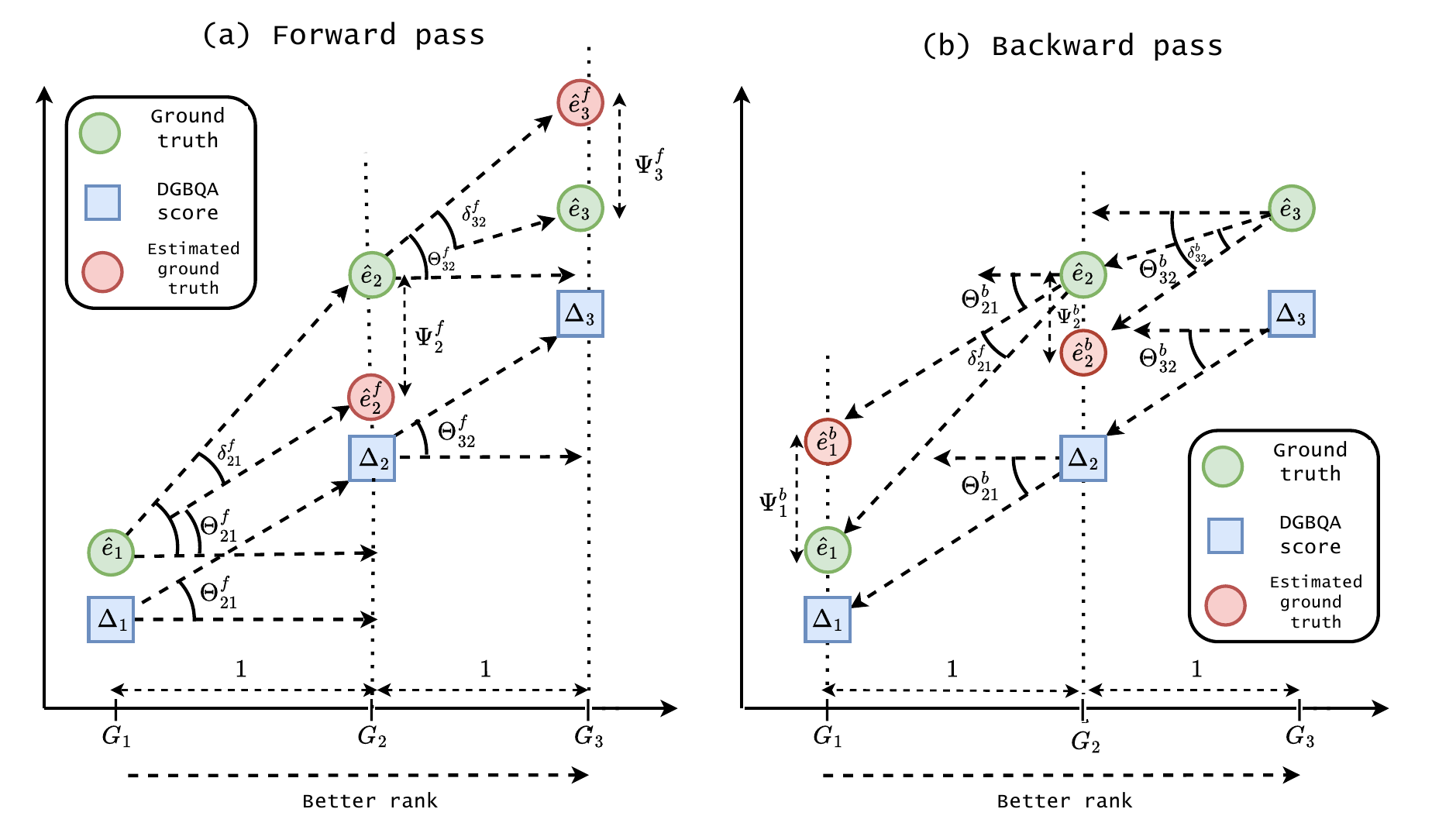}
    \caption{\textbf{Computation of trend match distance ($\Psi$}): With $\Psi$ we find out if DGBQA scores of two consecutively ranked gestures have the same biometric separation as the ground truth.
    Firstly, all the gestures are arranged as per the ranking order in ground truth. The last gesture (here $G_3$) carries the highest biometric characteristics. 
    To compute $\Psi$ we traverse in two directions: (i) forward pass: along increasing order (refer (a)) and (ii) backward pass: along decreasing order (refer (b)). We utilize the difference between consecutive DGBQA scores to estimate the ground truth. 
    The summation of difference between estimated and ground truth score from both the passes gives $\Psi$ for that gesture. 
    This is summed across all the gestures to derive $\Psi$.} 
    \label{fig:MS_PatternMatch}
\end{figure*}

\section{Proposed: Advanced acceptance score}
\label{sec:proposed}

\subsection{Background: The DGBQA framework}
\noindent The DGBQA framework \cite{verma2024quantifying} operates in two stages. 
For the first stage, a feature extractor $f_{\theta}(.)$ jointly generates an identity and gesture aware feature space. 
This is achieved by optimizing an multi-objective loss function:

\begin{equation}
\mathcal{L}_{Obj} = \mathcal{L}_{HGR} + \lambda_{ID}\mathcal{L}_{ID} + \lambda_{ICGD}\mathcal{L}_{ICGD}
\label{eq:complete_loss}
\end{equation}

Here, $\mathcal{L}_{HGR},~\mathcal{L}_{ID},~\text{and}~\mathcal{L}_{ICGD}$ respectively represent the gesture recognition, identity recognition, and identity-level cross-gesture disentanglement (ICGD) loss. Also, $\lambda_{ID}~$ and $\lambda_{ICGD}$ are the weighting parameters. 
The ICGD limits the entanglement of identity features between different gestures. Let there be $G$ gesture classes from $I$ identities. Also, let the batch size be $N$, and the final embeddings extracted from the feature extractor be given by: $\bar{f}~\in~\mathbb{R}^{N\times d}$. Then, $\mathcal{L}_{ICGD}$ is given by:

\begin{equation}
    \mathcal{L}_{ICGD} = \frac{1}{I} \mathlarger{\sum_{i=1}^{I}}\left(\frac{\mathlarger{\sum_{m=1}^N\sum_{n=1}^N} {(({\bar{f}}\bar{f}^T) \odot M_i)}_{mn}}{\mathlarger{\sum_{m=1}^N \sum_{n=1}^N} M_{i,mn} + 1}\right)
\label{eq:LICGD}
\end{equation}
\noindent Here, $M \in \mathbb{R}^{N\times N};~M_{m,n}\in \{0,1\},~\forall~m,n~\in~\{1,...,N\}$ is a mask. Any element $M_{m,n}$ from the mask is unity if (i) $m\neq n~$, (ii) ${\bar{f}_m}^T\bar{f}_n\geq 0$, and (iii) $~m^{th}~\text{and}~n^{th}$ samples belongs to different gesture but the same identity class 

After optimizing the $\mathcal{L}_{Obj}$, embeddings corresponding to the test set are extracted. The generated feature space comprises of gesture clusters, and within them smaller identity clusters. Then for each gesture cluster $j~\in~\{1,...,G\}$, two parameters are computed: (i) uniqueness ($d_{UNQ_{j}}$): average distance between the subject clusters) and variability. (ii) Variability ($d_{VRB_{j}}$): average variability in an identity clusters, averaged over all identities. 
These parameters are then used to compute the DGBQA score ($\Delta_{j};~\Delta~\in~\mathbb{R}^{G}$) for that gesture. 

\begin{equation}
\Delta_{GBQA_j} = \exp{(d_{UNQ_j}-d_{VRB_j})}-\left(\frac{d_{VRB_j}}{d_{UNQ_j}}\right)
\label{eq:DGBQA}
\end{equation}

\subsection{Preliminaries}
In order to evaluate DGBQA scores, we need ground truth. Thus, we resort upon equal error rates (EER). However, (i) EER is optimal when it is minimum, while the opposite is true for DGBQA scores. (ii) The scale of EER and DGBQA does not match. To address these, we consider $100-EER$ values. The resultant values from $G$ gestures are stacked in a vector. Then, we rescale them using z-score normalization followed by l2 normalization. The output $\hat{e}~\in~\mathbb{R}^G$ is used as the ground truth. The above discussed normalization scheme is also applied for the DGBQA scores. Thus, after normalization both DGBQA and ground truth are in range $\left[-1,1\right]$.

\subsection{Evaluation measures for design criteria}

\subsubsection{\textbf{Rank deviation}} We utilize the rank deviation parameter given by \cite{verma2024quantifying}. Firstly, both DGBQA and ground truth scores are sorted in descending order. For a gesture $j$, let the rank derived wrt to DGBQA and ground truth scores (after sorting) be given by $r^{\Delta}_{j}$ 
and $r^{\hat{e}}_{j}$. Mathematically, the rank deviation ($\hat{e}$) is given by:

\begin{equation}
    \hat{r}~=~\frac{1}{G}\left(\mathlarger{\sum_{j=1}^{G} \Arrowvert r_{j}^{\Delta} - r_{j}^{\hat{e}} \Arrowvert_{1}
    }\right)
\label{eq:rank_dev}
\end{equation}

\subsubsection{\textbf{Relevance}} Relevance parameter is rewarding if DGBQA scores are high for high ranking gestures, and low for low ranking gestures.  
To achieve this, we: (i) parametrize the relevance with rank and (ii) consider inverse of DGBQA scores. 
Firstly, we arrange DGBQA scores as per the ground truth ranking order. Let $\Delta \left[k\right]$ represent the DGBQA score of the $k^{\text{th}}$ gesture. Then, the relevance for a gesture $j$ is given by:

\begin{equation}
    \mathlarger{\mathcal{R}_{j}=\gamma\Bigl({\frac{G-{r_{j}^{\hat{e}}}+1}{G}\Bigr){\Delta\left[r_{j}^{\hat{e}}\right]}}+\Bigl(\frac{r_{j}^{\hat{e}}}{G}\Bigr)\Bigl(1-\Delta\left[r_{j}^{\hat{e}}\right]\Bigr)}
\label{eq:relevance_j}
\end{equation}

\noindent Here, $1-\Delta\left[r_{j}^{\hat{e}}\right]$ is the inverse of DGBQA score $\Delta\left[r_{j}^{\hat{e}}\right]$. Also, $\gamma$ is a weighting factor set to $\gamma=2$. Therefore, for the top most ranked gesture $r_{j}^{\hat{e}}=1$, the corresponding relevance value ($R_{1}$) is given by:

\begin{equation}
    \mathlarger{\mathcal{R}_{1}=\Delta \left[1\right]+\Bigl(\frac{1}{G}\Bigr)\Bigl(1-\Delta\left[1\right]\Bigr)}
\label{eq:relevance_high}
\end{equation}

\noindent While for the lowest ranked gesture $r_{j}^{\hat{e}}=G$, relevance takes the form:

\begin{equation}
    \mathlarger{\mathcal{R}_{G}=\Bigl(\frac{1}{G}\Bigr)\Delta \left[G\right]+\Bigl(1-\Delta\left[G\right]\Bigr)}
\label{eq:relevance_low}
\end{equation}

\noindent Thus, from eq. \ref{eq:relevance_high} and \ref{eq:relevance_low} one can conclude that the relevance term (eq. \ref{eq:relevance_j}) asserts more weightage to: (i) DGBQA scores for higher ranks and (ii) inverse of DGBQA scores, for the lower ranks. This rank adaptive behavior enables it to evaluate as per the desiderata. Thus, the overall relevance ($\mathcal{R}$) is given by:

\begin{equation}
    \mathlarger{\mathcal{R}=\sum_{j=1}^{G}\mathcal{R}_j}
\label{eq:relevance_total}
\end{equation}

\noindent Because of z-score normalization, low ranking DGBQA scores assumes negative values. This can lead to lower ranking gestures dominating the relevance. To prevent this, we increase the weightage of DGBQA score values (the first in eq. \ref{eq:relevance_j}) by weighting factor $\gamma$.

\label{sec:relevance}

\vspace{0.15cm}

\subsubsection{\textbf{Entanglement}} In order to measure residual entanglement, we utilize the ICGD score \cite{verma2024quantifying}. Higher the ICGD score, lower the reliability in the DGBQA score values. Mathematically,

\begin{equation}
    \mathlarger{C_d = \frac{1}{I} \mathlarger{\sum_{i=1}^{I}}\left(\frac{\mathlarger{\sum_{m=1}^N\sum_{n=1}^N} {(({\bar{f}}\bar{f}^T) \odot M_i)}_{mn}}{\mathlarger{\sum_{m=1}^N \sum_{n=1}^N} M_{i,mn}}\right)}
\label{eq:ICGD_Score}
\end{equation}

\vspace{0.15cm}

\subsubsection{\textbf{Trend deviation}}

For DGBQA scores to be optimal, we expect the change in consecutive score values to be similar to the one observed in ground truth scores. 
This ensures that physical significance of biometric characteristics present in the ground truth is also reflected in the DGBQA scores. 
To measure deviation from this property, we propose trend match distance ($\Psi$). 
We refer this as 'trend match' because it is equivalent to comparing the trends of score value progression. 

\vspace{0.15cm}

Fig. \ref{fig:MS_PatternMatch} depicts a representative example of $\Psi$ computation. We take a general case with $G$ gestures. We firstly arrange the DGBQA scores as per the ground truth ranking order. So, $j=G$ is the top most ranking gesture. As illustrated, there are two phases of computation: (i) forward pass and (ii) backward pass. The results from the both phases are finally aggregated to derive $\Psi$.

\noindent \textbf{Forward pass:} For any gesture $G_j~\&~j~\in~\{2,...,G\}$, we estimate the ground truth value ($\hat{e}_{j}$) using the slope between $\Delta_j~\&~\Delta_{j-1}$. Let the angle subtended between $\Delta_j~\&~\Delta_{j-1}$ be given by $\Theta_{j,j-1}^f$. Then, the prediction ($\hat{e}_{j}^f$) is given by:

\begin{align}
    \tan(\Theta_{j,j-1}^f)=\frac{\Delta_{j}-\Delta_{j-1}}{j-(j-1)}&=\Delta_{j}-\Delta_{j-1} \label{eq:fwd_slope} \\
    \hat{e}_{j}^f = e_{j-1}+\tan(\Theta_{j,j-1}^f)=e_{j-1}&+(\Delta_{j}-\Delta_{j-1}) \label{eq:fwd_pass}
\end{align}

\noindent Thus, the error between the ground truth and predicted value gives the forward pass error ($\Psi_{j}^{f}$). Mathematically,

\begin{equation}
    \Psi_{j}^{f}~=~|{e}_{j}^f - {e}_{j}|
    \label{eq:fwd_error}
\end{equation}

\vspace{0.15cm}


\noindent \textbf{Backward pass:} As illustrated in Fig. \ref{fig:MS_PatternMatch}, in backward pass, we traverse the DGBQA scores in descending order. Similar to forward pass, for any gesture $G_j~\&~j~\in~\{1,...,G-1\}$, we estimate the ground truth value via the slope between $\Delta_{j+1}~\&~\Delta{j}$. 
Let the angle subtended by $\Delta_j$ on $\Delta_{j+1}$ be given by $\Theta_{j+1,j}^b$. Thus, the slope is given by $\tan(\pi+\Theta_{j+1,j}^b)$. Mathematically,

\begin{equation}
\tan(\pi+\Theta_{j+1,j}^b)=\frac{\Delta_{j+1}-\Delta{j}}{(j+1)-j} = \Delta_{j+1}-\Delta{j}
\label{eq:back_slope}
\end{equation}

\noindent From the slope, we estimate ground truth ($\hat{e}_{j}^b$) value:

\begin{equation}
\hat{e}_{j}^b = e_{j+1}-\tan(\Theta_{j+1,j}^b)=e_{j+1}-(\Delta_{j+1}-\Delta_{j})
\label{eq:back_estimate}
\end{equation}

\noindent Therefore, the error in the backward pass 
is given by:

\begin{equation}
    \Psi_{j}^{b}~=~|{e}_{j}^b - {e}_{j}|
\label{eq:back_error}
\end{equation}

\noindent For a gesture, the total error is cumulative of errors in both forward and backward pass. This ensures that trend in both the direction is observed. $\Psi$ is aggregation of errors for all the gestures, thus:

\begin{equation}
   \Psi=\frac{\bigl[\sum_{j=2}^{G-1}\bigl(\Psi_{j}^f+\Psi_{j}^b\bigr)\bigr]~+~2(\Psi_{G}^f+\Psi_{1}^b)}{2}
\label{eq:trend_match}
\end{equation}

\noindent Thus, eq. \ref{eq:trend_match} metricizes the amount of error in predicting ground truth from DGBQA score progressions. 
Except $j=1~\&~j=G$, errors for all the other gestures will be computed in both forward and backward pass. To compensate this, we double their errors during computation.

\begin{table*}[!htb]
\tiny
\centering
\caption{\textbf{Comparison of advanced acceptance score with it's variants}. We compare the score selected as optimal by $A_r^{*}(\Delta)$ and its variants: (i) individual measures and (ii) their different combinations. We assess on the basis of design criterion. The proposed $A_r^{*}(\Delta)$ outperforms the other variants on all three datasets. The selected score are jointly optimal in terms of design criterion.}
\label{tab:compVariants}
\begin{tabular}{|c|ccccc|ccccc|ccccc|}
\hline
\multirow{2}{*}{\textbf{\begin{tabular}[c]{@{}c@{}}Evaluation\\ Measure\end{tabular}}} & \multicolumn{5}{c|}{\textbf{Soli}}                                                                                                                                                     & \multicolumn{5}{c|}{\textbf{HandLogin}}                                                                                                                                               & \multicolumn{5}{c|}{\textbf{TinyRadar}}                                                                                                                                                \\ \cline{2-16} 
                                                                                       & \multicolumn{1}{c|}{\textbf{Model}}  & \multicolumn{1}{c|}{\textbf{$\hat{r}$}} & \multicolumn{1}{c|}{\textbf{$\mathcal{R}$}} & \multicolumn{1}{c|}{\textbf{$\Psi$}} & \textbf{$C_{d}$} & \multicolumn{1}{c|}{\textbf{Model}} & \multicolumn{1}{c|}{\textbf{$\hat{r}$}} & \multicolumn{1}{c|}{\textbf{$\mathcal{R}$}} & \multicolumn{1}{c|}{\textbf{$\Psi$}} & \textbf{$C_{d}$} & \multicolumn{1}{c|}{\textbf{Model}}  & \multicolumn{1}{c|}{\textbf{$\hat{r}$}} & \multicolumn{1}{c|}{\textbf{$\mathcal{R}$}} & \multicolumn{1}{c|}{\textbf{$\Psi$}} & \textbf{$C_{d}$} \\ \hline
\textbf{$\hat{r}$}                                                                     & \multicolumn{1}{c|}{ViViT (1.5,0.5)} & \multicolumn{1}{c|}{0.45}               & \multicolumn{1}{c|}{8.54}                   & \multicolumn{1}{c|}{1.70}            & 0.351            & \multicolumn{1}{c|}{MViT (0.5,1.0)} & \multicolumn{1}{c|}{0.00}               & \multicolumn{1}{c|}{4.11}                   & \multicolumn{1}{c|}{2.09}            & 0.327            & \multicolumn{1}{c|}{ViViT (0.5,1.5)} & \multicolumn{1}{c|}{1.27}               & \multicolumn{1}{c|}{8.39}                   & \multicolumn{1}{c|}{2.24}            & 0.488            \\ \hline
\textbf{$\mathcal{R}$}                                                                 & \multicolumn{1}{c|}{MF (1.5,0.5)}    & \multicolumn{1}{c|}{0.45}               & \multicolumn{1}{c|}{8.71}                   & \multicolumn{1}{c|}{1.56}            & 0.361            & \multicolumn{1}{c|}{MViT (1.0,1.5)} & \multicolumn{1}{c|}{0.00}               & \multicolumn{1}{c|}{4.12}                   & \multicolumn{1}{c|}{1.25}            & 0.284            & \multicolumn{1}{c|}{ViViT (1.0,2.5)} & \multicolumn{1}{c|}{1.45}               & \multicolumn{1}{c|}{8.50}                   & \multicolumn{1}{c|}{2.13}            & 0.540            \\ \hline
\textbf{$\Psi$}                                                                        & \multicolumn{1}{c|}{ViViT (1.0,0.5)} & \multicolumn{1}{c|}{1.54}               & \multicolumn{1}{c|}{8.32}                   & \multicolumn{1}{c|}{1.36}            & 0.309            & \multicolumn{1}{c|}{MViT (1.0,1.0)} & \multicolumn{1}{c|}{0.50}               & \multicolumn{1}{c|}{3.95}                   & \multicolumn{1}{c|}{0.82}            & 0.373            & \multicolumn{1}{c|}{TAM (1.0,1.0)}   & \multicolumn{1}{c|}{2.00}               & \multicolumn{1}{c|}{8.00}                   & \multicolumn{1}{c|}{1.99}            & 0.651            \\ \hline
\textbf{$C_{d}$}                                                                       & \multicolumn{1}{c|}{TPN (1.0,0.5)}   & \multicolumn{1}{c|}{1.72}               & \multicolumn{1}{c|}{8.12}                   & \multicolumn{1}{c|}{2.39}            & 0.088            & \multicolumn{1}{c|}{TPN (1.5,2.5)}  & \multicolumn{1}{c|}{1.25}               & \multicolumn{1}{c|}{1.42}                   & \multicolumn{1}{c|}{3.95}            & 0.001            & \multicolumn{1}{c|}{ViViT (0.5,2.5)} & \multicolumn{1}{c|}{1.27}               & \multicolumn{1}{c|}{8.34}                   & \multicolumn{1}{c|}{2.59}            & 0.469            \\ \hline
\textbf{$A_r(\Delta)$~\cite{verma2024quantifying}}                                                                 & \multicolumn{1}{c|}{ViViT (0.5,1.0)} & \multicolumn{1}{c|}{0.45}               & \multicolumn{1}{c|}{8.54}                   & \multicolumn{1}{c|}{1.70}            & 0.351            & \multicolumn{1}{c|}{MViT (0.5,1.0)} & \multicolumn{1}{c|}{0.00}               & \multicolumn{1}{c|}{4.11}                   & \multicolumn{1}{c|}{2.09}            & 0.327            & \multicolumn{1}{c|}{MF (1.0,1.5)}    & \multicolumn{1}{c|}{1.45}               & \multicolumn{1}{c|}{8.28}                   & \multicolumn{1}{c|}{2.27}            & 0.484            \\ \hline
\textbf{$A_r(\Delta)*\bar{C_{d}}$}                                                     & \multicolumn{1}{c|}{MViT (0.5,1.0)}  & \multicolumn{1}{c|}{0.81}               & \multicolumn{1}{c|}{8.69}                   & \multicolumn{1}{c|}{1.63}            & 0.123            & \multicolumn{1}{c|}{MViT (1.0,1.5)} & \multicolumn{1}{c|}{0.00}               & \multicolumn{1}{c|}{4.12}                   & \multicolumn{1}{c|}{1.25}            & 0.284            & \multicolumn{1}{c|}{MF (1.0,1.5)}    & \multicolumn{1}{c|}{1.45}               & \multicolumn{1}{c|}{8.28}                   & \multicolumn{1}{c|}{2.27}            & 0.484            \\ \hline
\textbf{$A_r(\Delta)*\bar{\Psi}$}                                                      & \multicolumn{1}{c|}{ViViT (0.5,1.0)} & \multicolumn{1}{c|}{0.45}               & \multicolumn{1}{c|}{8.54}                   & \multicolumn{1}{c|}{1.70}            & 0.351            & \multicolumn{1}{c|}{MViT (1.0,1.5)} & \multicolumn{1}{c|}{0.00}               & \multicolumn{1}{c|}{4.12}                   & \multicolumn{1}{c|}{1.25}            & 0.284            & \multicolumn{1}{c|}{MF (1.0,1.5)}    & \multicolumn{1}{c|}{1.45}               & \multicolumn{1}{c|}{8.28}                   & \multicolumn{1}{c|}{2.27}            & 0.484            \\ \hline
\textbf{$\bar{\Psi}*\bar{C_d}$}                                                        & \multicolumn{1}{c|}{MViT (0.5,1.0)}  & \multicolumn{1}{c|}{0.81}               & \multicolumn{1}{c|}{8.69}                   & \multicolumn{1}{c|}{1.63}            & 0.123            & \multicolumn{1}{c|}{TPN (1.5,1.0)}  & \multicolumn{1}{c|}{0.5}                & \multicolumn{1}{c|}{4.02}                   & \multicolumn{1}{c|}{2.13}            & 0.009            & \multicolumn{1}{c|}{MF (1.0,1.5)}    & \multicolumn{1}{c|}{1.45}               & \multicolumn{1}{c|}{8.28}                   & \multicolumn{1}{c|}{2.27}            & 0.484            \\ \hline
\textbf{$\mathcal{R}*\bar{C_d}$}                                                      & \multicolumn{1}{c|}{MViT (0.5,1.0)}  & \multicolumn{1}{c|}{0.81}               & \multicolumn{1}{c|}{8.69}                   & \multicolumn{1}{c|}{1.63}            & 0.123            & \multicolumn{1}{c|}{TPN (1.5,1.0)}  & \multicolumn{1}{c|}{0.5}                & \multicolumn{1}{c|}{4.02}                   & \multicolumn{1}{c|}{2.13}            & 0.009            & \multicolumn{1}{c|}{ViViT (0.5,2.5)} & \multicolumn{1}{c|}{1.27}               & \multicolumn{1}{c|}{8.34}                   & \multicolumn{1}{c|}{2.59}            & 0.469            \\ \hline
\textbf{\begin{tabular}[c]{@{}c@{}}$nA_r^{*}(\Delta)$ \\ (Proposed)\end{tabular}}      & \multicolumn{1}{c|}{MViT (0.5,1.0)}  & \multicolumn{1}{c|}{0.81}               & \multicolumn{1}{c|}{8.69}                   & \multicolumn{1}{c|}{1.63}            & 0.123            & \multicolumn{1}{c|}{MViT (1.0,1.5)} & \multicolumn{1}{c|}{0.00}               & \multicolumn{1}{c|}{4.12}                   & \multicolumn{1}{c|}{1.25}            & 0.284            & \multicolumn{1}{c|}{MF (1.0,1.5)}    & \multicolumn{1}{c|}{1.45}               & \multicolumn{1}{c|}{8.28}                   & \multicolumn{1}{c|}{2.27}            & 0.484            \\ \hline
\end{tabular}
\end{table*}

\begin{figure*}[]
    \centering
    \includegraphics[scale=0.73]{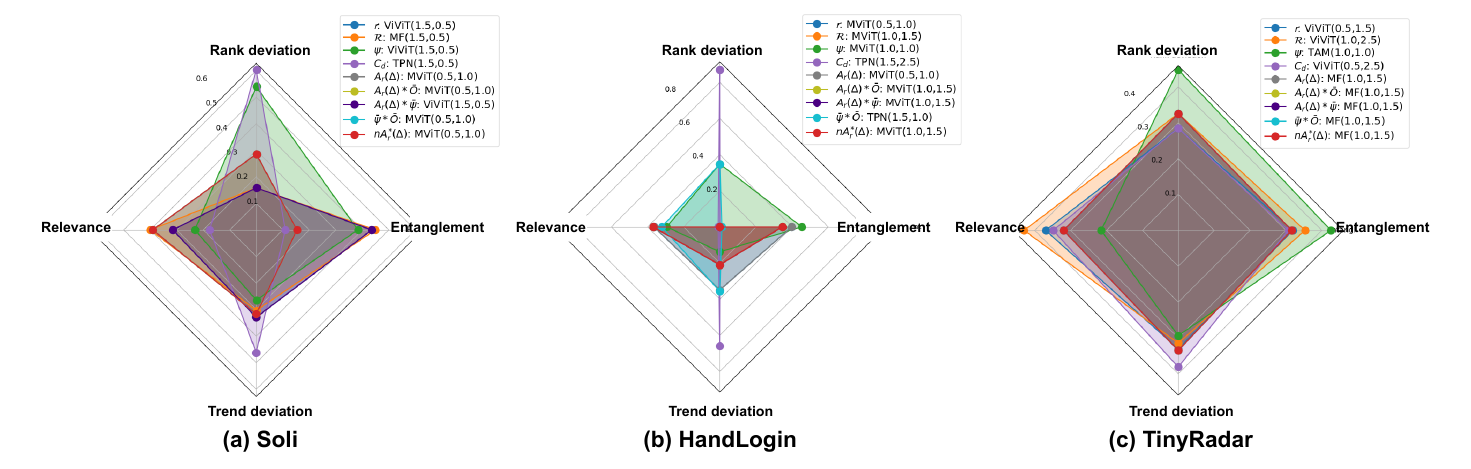}
    \caption{\textbf{Radar charts for comparing score selection of advanced acceptance score and it's variants.} Comparison in performed in terms of our design criterion: (i) rank deviation, (ii) relevance, (iii) trend deviation, and (iv) entanglement. To clearly discriminate between the relevance scores, we raised them as an exponent: $2^{\lambda\mathcal{R}}$. In comparison to it's variants, the advanced acceptance score satisfies all the criteria. This high end capabilities can be accredited to multi-faceted evaluation in the proposed. Exact numerical values can be inferred from Table \ref{tab:compVariants}, while we illustrate a qualitative comparison in Fig. \ref{fig:MS_QA}.}
    \label{fig:MS_Inter_Radar}
\end{figure*}

\subsection{Advanced acceptance score}
Verma et al. \cite{verma2024quantifying} gave the acceptance score ($A_r(.)$). For DGBQA score $\Delta$, we have:

\begin{equation}
A_r(\Delta)=\sum_{j=1}^{G}\frac{2^{\lambda \Bigl[ \gamma\Bigl({\frac{G-{r_{j}^{\hat{e}}}+1}{G}\Bigr){\Delta\left[r_{j}^{\hat{e}}\right]}}+\Bigl(\frac{r_{j}^{\hat{e}}}{G}\Bigr)\Bigl(1-\Delta\left[r_{j}^{\hat{e}}\right]\Bigr) \Bigr]}}{\Arrowvert r_{j}^{\Delta} - r_{j}^{\hat{e}} \Arrowvert_{1}}
\label{eq:Ar}
\end{equation}

 $A_r(\Delta)$ neither evaluates for entanglement, nor for trend deviation. 
Thus, to achieve holistic evaluation, we unify the above discussed measures into the advanced acceptance score ($A_r^*(.)$). We define $A_r^*(.)$ as:

\begin{align}
     &A_r^*(\Delta)=\frac{\mathlarger{\sum_{j=1}^{G}}\Bigl( \frac{2^{\lambda\mathcal{R}_j}}{\exp(\kappa*\Arrowvert r_{j}^{\Delta} - r_{j}^{\hat{e}} \Arrowvert_{1})}\Bigr)}{\sqrt{\log_2(2+\nu\Psi)}}*{\exp(-\beta C_d)}
\label{eq:Ar_star}
\end{align}

\noindent Here, $\lambda,~\kappa,~\nu,~\text{and}~\beta$ are scaling factors.  
Rank deviation and entanglement are higher in priority, hence we assign them more weights. 
This is because we at least require the ranking order to be correct and reliable. 
To facilitate comparison between results on different datasets an, we further normalize $A_r^*(.)$. The resultant $nA_r^*(.)$ is defined as:

\begin{equation}
nA_r^*(\Delta)=\frac{A_r^{*}(\Delta)}{A_r^{*}(\hat{e})}
\label{eq:nAr_star}     
\end{equation}
\vspace{-0.15cm}
\noindent Where,
\vspace{-0.15cm}
\begin{equation}
    A_r^*(\hat{e})=\mathlarger{\sum_{j=1}^{G}}2^{\lambda \Bigl[ \gamma\Bigl({\frac{G-{r_{j}^{\hat{e}}}+1}{G}\Bigr){\hat{e}\left[r_{j}^{\hat{e}}\right]}}+\Bigl(\frac{r_{j}^{\hat{e}}}{G}\Bigr)\Bigl(1-\hat{e}\left[r_{j}^{\hat{e}}\right]\Bigr) \Bigr]}
\label{eq:Ar_star_e}    
\end{equation}

\section{Experimental analysis}
\label{sec:experimetal}

\subsection{Architecture used}
For extracting DGBQA scores we utilized five deep learning architectures. 
Each of them comprises of an initial 3D CNN backbone (as used in \cite{verma2024quantifying, jaswal2021range}), followed by a state-of-the-art spatiotemporal modeling network: 
(i) ViViT \cite{arnab2021vivit}, 
(ii) MotionFormer (MF) \cite{patrick2021keeping}, 
(iii) Multiscale Vision Transformer (MViT) \cite{fan2021multiscale}, 
(iv) Temporal pyramid network (TPN) \cite{yang2020temporal}, 
and Temporal adaptive module (TAM) \cite{liu2021temporal}.
The naming convention of these architectures are: '$<\text{abbreviation of the architecture}>(\lambda_{ID},\lambda_{ICGD})$'.

\subsection{Experiment design and protocol}
We train different feature extractors by varying the architecture and hyperparmeter settings: $\lambda_{ID}=\{0.5,1.0,1.5\}$ and $\lambda_{ICGD}=\{0.5,1.0,1.5,2.5\}$. 
Then, we compare the DGBQA scores of these feature extractors based on evaluation measures. 
The DGBQA scores designated optimal by the measures are further compared on the basis of four design criterion.
%
In order to validate the effectivity of the proposed, we perform this comparison in two settings: (i) $A_r^*(.)$ versus it's variants and (ii) $A_r^*(.)$ versus existing measures. 
It must be noted that at intra dataset level, both  $nA_r^*(.)$ and  $A_r^*(.)$ results in same selection. In our experiments we use, $k=1,~\beta=0.75,~\lambda=2,~\text{and}~\nu=1$. However, we also report the impact of changing the hyperparameters on $nA_r^*(.)$ values. 
For ground truth values, we use the ones reported by \cite{verma2024quantifying}. For all the experiments we split the dataset in the ratio of $60:40$ for training and testing.

\subsection{Dataset used}
\noindent Following are the publicly available datasets used for conducting experiments: 

\noindent \textbf{1) Soli} \cite{wang2016interacting}: This dataset contains a total of $2750$ range-Doppler image sequences of hand gestures. It has been collected for $11$ gestures which has been performed by $10$ identities. 

\noindent \textbf{2) HandLogin} \cite{wu2015leveraging}: This dataset comprises of depth-maps for $4$ gestures with $15$ subjects. 
Similar to \cite{verma2024quantifying} we generate first-temporal difference map for the sequences~\cite{sheng2022progressive}.
This ensures that the biometric scores are derived only from motion details and do not consider any physiological details. It is to be noted that, in contrast to RGB and depth videos, range-Doppler sequences does not posses explicit physiological details. 

\noindent \textbf{3) TinyRadar} \cite{scherer2021tinyradarnn}: This is a large-scale range-Doppler hand gesture recognition dataset with $30,300$ samples. 
It comprises of the same $11$ gesture as Soli. However, it has been collected using a different radar sensor and that too from a larger identity base of $26$ subjects. 
Please note, the samples of TinyRadar occupy larger dimensions. To meet with GPU memory constraints we had to decrease the batch size. However, a larger number of identities in the dataset prohibited disentanglement pairs to be formed. This resulted is inferior disentanglement in comparison to the other datasets.

\subsection{Comparison of advanced acceptance score with it's variants}
We compare the model selection by the proposed and it's variants: (i) individual evaluation measures and (ii) their combination. 
For combining $C_d$ and $\Psi$, we utilize their penalty forms, $\bar{C_d}=\exp(-\beta*C_d)$ and $\bar{\Psi}=1/\sqrt{\log_2(2+\nu\Psi)}$.
We perform comparison on two levels. Firstly, we seek the difference in model selection and then compare the distribution of evaluation measures.

\begin{figure*}[!t]
    \centering
    \includegraphics[width=1.0\textwidth]{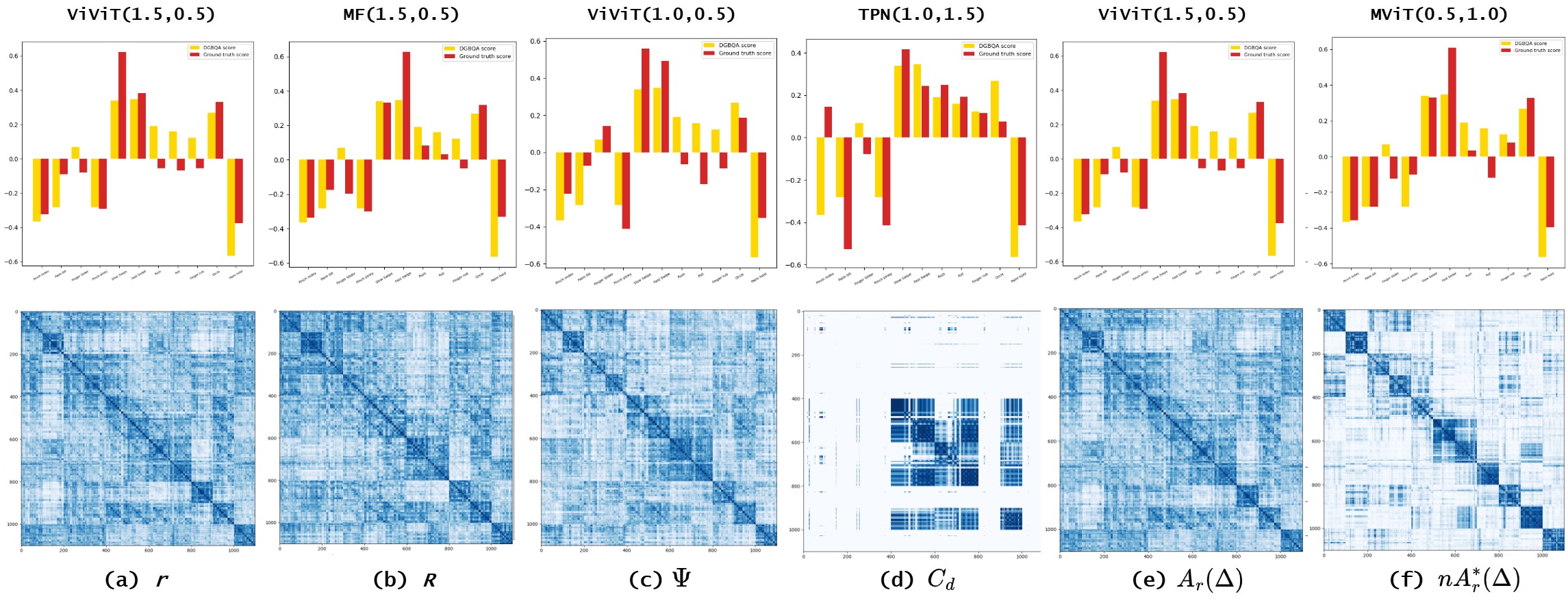}
    \caption{\textbf{Qualitative comparison between advanced acceptance score and its variants over score/model selection} (on Soli dataset). We conduct this analysis using: (i) histogram comparing selected DGBQA scores and ground truth (see the first row) and (ii) Gram matrix (to quantify entanglement). The score/model selected from $A_r^*(\Delta)$ (column (f)) has DGBQA scores comparable to the ground truth. While it attains the least entanglement (we require lighter values in off diagonal elements of the off diagonal blocks). This is also observed wrt \cite{verma2024quantifying}.}
    \label{fig:MS_QA}
\end{figure*}

\begin{figure}[]
    \centering
    \includegraphics[width=1.0\linewidth]{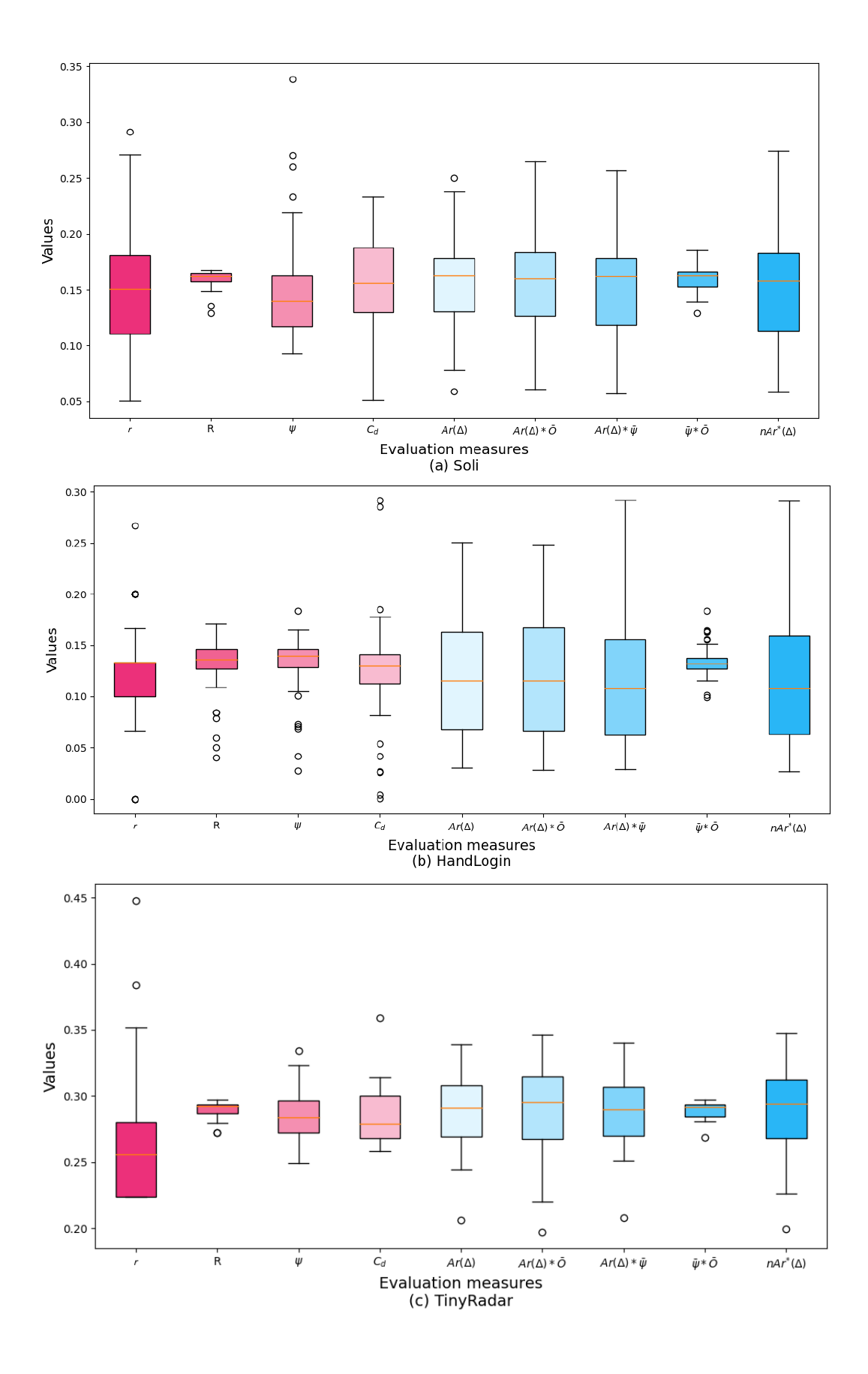}
    \caption{\textbf{Bars and whiskers plot to compare distribution of evaluation measures.} The plot highlights that individual measures covers different range. This suggests that they convey different information. $A_r^{*}(\Delta)$ fuses them to generate a broader range of values. This is pivotal in differentiating between good and low quality scores.}
    \label{fig:MS_BarWhiskers}
\end{figure}

\begin{table*}[!htb]
\tiny
\centering
\caption{\textbf{Comparison of the proposed with SOTA in terms of score selection.} To conduct this study, we adapted retrieval quality measures \cite{jarvelin2002cumulated, chapelle2009expected, sakai2013summaries, lioma2017evaluation, yilmaz2008estimating, gienapp2020impact, diaz2022offline} for our problem. Unlike existing measures, the score selected by $A_r^*(\Delta)$ considers entanglement and trend deviation. This allows it jointly satisfy all the design criteria. For fair comparison, we also our quality measure: relevance ($\mathcal{R}$) in comparison.
It is evident from the results, that $\mathcal{R}$ consistently outperformed the other measures on the three datasets over rank deviation, relevance, and trend deviation. This can be accredited to its ability to reward both high and low ranking gestures as per their score values. A pictorial representation of this analysis has been illustrated with radar charts in Fig. \ref{fig:MS_Comp_Radar}.
}
\label{tab:compSOTA}
\begin{tabular}{|c|ccccc|ccccc|ccccc|}
\hline
                                                                                        & \multicolumn{5}{c|}{\textbf{Soli}}                                                                                                                                                     & \multicolumn{5}{c|}{\textbf{HandLogin}}                                                                                                                                               & \multicolumn{5}{c|}{\textbf{TinyRadar}}                                                                                                                                                \\ \cline{2-16} 
\multirow{-2}{*}{\textbf{\begin{tabular}[c]{@{}c@{}}Evaluation\\ Measure\end{tabular}}} & \multicolumn{1}{c|}{\textbf{Model}}  & \multicolumn{1}{c|}{\textbf{$\hat{r}$}} & \multicolumn{1}{c|}{\textbf{$\mathcal{R}$}} & \multicolumn{1}{c|}{\textbf{$\Psi$}} & \textbf{$C_{d}$} & \multicolumn{1}{c|}{\textbf{Model}} & \multicolumn{1}{c|}{\textbf{$\hat{r}$}} & \multicolumn{1}{c|}{\textbf{$\mathcal{R}$}} & \multicolumn{1}{c|}{\textbf{$\Psi$}} & \textbf{$C_{d}$} & \multicolumn{1}{c|}{\textbf{Model}}  & \multicolumn{1}{c|}{\textbf{$\hat{r}$}} & \multicolumn{1}{c|}{\textbf{$\mathcal{R}$}} & \multicolumn{1}{c|}{\textbf{$\Psi$}} & \textbf{$C_{d}$} \\ \hline
\textbf{RMSE}                                                                           & \multicolumn{1}{c|}{TPN (1.5,1.0)}   & \multicolumn{1}{c|}{1.00}               & \multicolumn{1}{c|}{8.62}                   & \multicolumn{1}{c|}{1.39}            & 0.331           & \multicolumn{1}{c|}{MViT (1.0,1.0)} & \multicolumn{1}{c|}{0.50}               & \multicolumn{1}{c|}{3.95}                   & \multicolumn{1}{c|}{0.82}            & 0.372            & \multicolumn{1}{c|}{ViViT (1.0,2.5)} & \multicolumn{1}{c|}{1.45}               & \multicolumn{1}{c|}{8.50}                   & \multicolumn{1}{c|}{2.13}            & 0.540            \\ \hline
{\color[HTML]{000000} \textbf{$cos$}}                                                   & \multicolumn{1}{c|}{TPN (1.5,1.0)}   & \multicolumn{1}{c|}{1.00}               & \multicolumn{1}{c|}{8.62}                   & \multicolumn{1}{c|}{1.39}            & 0.331           & \multicolumn{1}{c|}{MViT (1.0,1.0)} & \multicolumn{1}{c|}{0.50}               & \multicolumn{1}{c|}{3.95}                   & \multicolumn{1}{c|}{0.82}            & 0.372            & \multicolumn{1}{c|}{ViViT (1.0,2.5)} & \multicolumn{1}{c|}{1.45}               & \multicolumn{1}{c|}{8.50}                   & \multicolumn{1}{c|}{2.13}            & 0.540            \\ \hline
\textbf{DCG~\cite{jarvelin2002cumulated}}                                                                            & \multicolumn{1}{c|}{ViViT (1.0,1.0)} & \multicolumn{1}{c|}{1.00}               & \multicolumn{1}{c|}{8.40}                   & \multicolumn{1}{c|}{1.65}            & 0.266            & \multicolumn{1}{c|}{MViT (0.5,0.5)} & \multicolumn{1}{c|}{0.50}               & \multicolumn{1}{c|}{3.93}                   & \multicolumn{1}{c|}{3.19}            & 0.313            & \multicolumn{1}{c|}{ViViT (0.5,1.0)} & \multicolumn{1}{c|}{1.45}               & \multicolumn{1}{c|}{8.44}                   & \multicolumn{1}{c|}{2.26}            & 0.502            \\ \hline
\textbf{$\tau$}                                                                         & \multicolumn{1}{c|}{ViViT (1.5,1.0)} & \multicolumn{1}{c|}{2.09}               & \multicolumn{1}{c|}{8.05}                   & \multicolumn{1}{c|}{1.76}            & 0.315            & \multicolumn{1}{c|}{TAM (1.0,2.5)}  & \multicolumn{1}{c|}{1.0}                & \multicolumn{1}{c|}{3.47}                   & \multicolumn{1}{c|}{3.60}            & 0.091            & \multicolumn{1}{c|}{ViViT (0.5,1.0)} & \multicolumn{1}{c|}{1.45}               & \multicolumn{1}{c|}{8.44}                   & \multicolumn{1}{c|}{2.26}            & 0.502            \\ \hline
\textbf{ERR \cite{chapelle2009expected}}                                                                            & \multicolumn{1}{c|}{MF (1.0,1.0)}    & \multicolumn{1}{c|}{1.00}               & \multicolumn{1}{c|}{8.64}                   & \multicolumn{1}{c|}{1.78}            & 0.268            & \multicolumn{1}{c|}{MViT (0.5,0.5)} & \multicolumn{1}{c|}{0.50}               & \multicolumn{1}{c|}{3.93}                   & \multicolumn{1}{c|}{3.19}            & 0.313            & \multicolumn{1}{c|}{ViViT (0.5,1.0)} & \multicolumn{1}{c|}{1.45}               & \multicolumn{1}{c|}{8.44}                   & \multicolumn{1}{c|}{2.26}            & 0.502            \\ \hline
\textbf{U-measure \cite{sakai2013summaries}}                                                                      & \multicolumn{1}{c|}{MF (1.0,1.5)}    & \multicolumn{1}{c|}{1.00}               & \multicolumn{1}{c|}{8.54}                   & \multicolumn{1}{c|}{1.84}            & 0.229            & \multicolumn{1}{c|}{MViT (1.0,1.5)} & \multicolumn{1}{c|}{0.00}               & \multicolumn{1}{c|}{4.12}                   & \multicolumn{1}{c|}{1.25}            & 0.284            & \multicolumn{1}{c|}{MF (1.0,1.0)}    & \multicolumn{1}{c|}{1.45}               & \multicolumn{1}{c|}{8.37}                   & \multicolumn{1}{c|}{2.27}            & 0.501            \\ \hline
\textbf{GRE \cite{lioma2017evaluation}}                                                                            & \multicolumn{1}{c|}{MF (1.5,0.5)}    & \multicolumn{1}{c|}{0.45}               & \multicolumn{1}{c|}{8.71}                   & \multicolumn{1}{c|}{1.56}            & 0.361            & \multicolumn{1}{c|}{MViT (0.5,1.0)} & \multicolumn{1}{c|}{0.00}               & \multicolumn{1}{c|}{4.11}                   & \multicolumn{1}{c|}{2.09}            & 0.327            & \multicolumn{1}{c|}{MF (1.0,1.5)}    & \multicolumn{1}{c|}{1.45}               & \multicolumn{1}{c|}{8.28}                   & \multicolumn{1}{c|}{2.27}            & 0.484            \\ \hline
\textbf{InfAp \cite{yilmaz2008estimating}}                                                                          & \multicolumn{1}{c|}{ViViT (1.5,0.5)} & \multicolumn{1}{c|}{0.45}               & \multicolumn{1}{c|}{8.71}                   & \multicolumn{1}{c|}{1.56}            & 0.361            & \multicolumn{1}{c|}{MViT (0.5,1.0)} & \multicolumn{1}{c|}{0.00}               & \multicolumn{1}{c|}{4.11}                   & \multicolumn{1}{c|}{2.09}            & 0.327            & \multicolumn{1}{c|}{ViViT (1.0,2.5)} & \multicolumn{1}{c|}{1.45}               & \multicolumn{1}{c|}{8.50}                   & \multicolumn{1}{c|}{2.13}            & 0.540            \\ \hline
\textbf{\begin{tabular}[c]{@{}c@{}}Negative\\ Relevance \cite{gienapp2020impact} \end{tabular}}                   & \multicolumn{1}{c|}{ViViT (1.0,1.5)} & \multicolumn{1}{c|}{1.90}               & \multicolumn{1}{c|}{8.01}                   & \multicolumn{1}{c|}{1.98}            & 0.217            & \multicolumn{1}{c|}{MF (1.5,1.5)}   & \multicolumn{1}{c|}{1.00}               & \multicolumn{1}{c|}{3.16}                   & \multicolumn{1}{c|}{4.18}            & 0.284            & \multicolumn{1}{c|}{ViViT (0.5,1.0)} & \multicolumn{1}{c|}{1.45}               & \multicolumn{1}{c|}{8.44}                   & \multicolumn{1}{c|}{2.26}            & 0.502            \\ \hline
\textbf{RPP \cite{diaz2022offline}}                                                                            & \multicolumn{1}{c|}{MViT (0.5,1.0)}  & \multicolumn{1}{c|}{0.81}               & \multicolumn{1}{c|}{8.69}                   & \multicolumn{1}{c|}{1.63}            & 0.123            & \multicolumn{1}{c|}{MViT (1.5,1.0)} & \multicolumn{1}{c|}{0.0}                & \multicolumn{1}{c|}{4.08}                   & \multicolumn{1}{c|}{2.20}            & 0.319            & \multicolumn{1}{c|}{ViViT (0.5,2.5)} & \multicolumn{1}{c|}{1.27}               & \multicolumn{1}{c|}{8.34}                   & \multicolumn{1}{c|}{2.59}            & 0.469            \\ \hline
\textbf{$A_r(\Delta)$~\cite{verma2024quantifying}}                                                                  & \multicolumn{1}{c|}{ViViT (0.5,1.0)} & \multicolumn{1}{c|}{0.45}               & \multicolumn{1}{c|}{8.54}                   & \multicolumn{1}{c|}{1.70}            & 0.351            & \multicolumn{1}{c|}{MViT (0.5,1.0)} & \multicolumn{1}{c|}{0.00}               & \multicolumn{1}{c|}{4.11}                   & \multicolumn{1}{c|}{2.09}            & 0.327            & \multicolumn{1}{c|}{MF (1.0,1.5)}    & \multicolumn{1}{c|}{1.45}               & \multicolumn{1}{c|}{8.28}                   & \multicolumn{1}{c|}{2.27}            & 0.484            \\ \hline
\textbf{\begin{tabular}[c]{@{}c@{}}$\mathcal{R}$\\ (Proposed)\end{tabular}}             & \multicolumn{1}{c|}{MF (1.5,0.5)}    & \multicolumn{1}{c|}{0.45}               & \multicolumn{1}{c|}{8.71}                   & \multicolumn{1}{c|}{1.56}            & 0.361            & \multicolumn{1}{c|}{MViT (1.0,1.5)} & \multicolumn{1}{c|}{0.00}               & \multicolumn{1}{c|}{4.12}                   & \multicolumn{1}{c|}{1.25}            & 0.284            & \multicolumn{1}{c|}{ViViT (1.0,2.5)} & \multicolumn{1}{c|}{1.45}               & \multicolumn{1}{c|}{8.50}                   & \multicolumn{1}{c|}{2.13}            & 0.540            \\ \hline
\textbf{\begin{tabular}[c]{@{}c@{}}$nA_r^{*}(\Delta)$ \\ (Proposed)\end{tabular}}       & \multicolumn{1}{c|}{MViT (0.5,1.0)}  & \multicolumn{1}{c|}{0.81}               & \multicolumn{1}{c|}{8.69}                   & \multicolumn{1}{c|}{1.63}            & 0.123            & \multicolumn{1}{c|}{MViT (1.0,1.5)} & \multicolumn{1}{c|}{0.00}               & \multicolumn{1}{c|}{4.12}                   & \multicolumn{1}{c|}{1.25}            & 0.284            & \multicolumn{1}{c|}{MF (1.0,1.5)}    & \multicolumn{1}{c|}{1.45}               & \multicolumn{1}{c|}{8.28}                   & \multicolumn{1}{c|}{2.27}            & 0.484            \\ \hline
\end{tabular}
\end{table*}

\begin{figure*}[]
    \centering
    \includegraphics[width=1.0\textwidth]{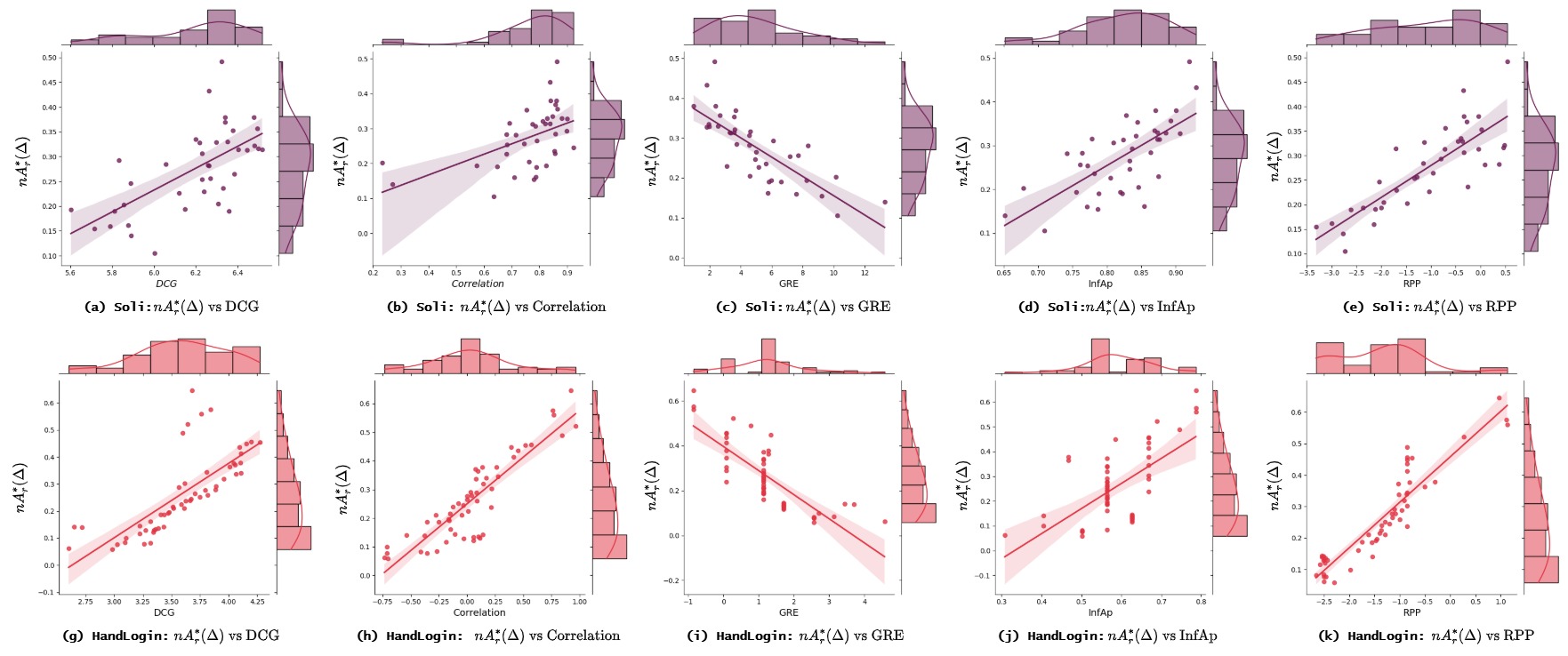}
    \caption{\textbf{Correlation analysis between the advanced acceptance score and the existing measures.} The figure shows that the proposed measure correlates well with existing measures. 
    This proves that it is both reliable and robust. 
    }
    \label{fig:Correlation}
\end{figure*}

\subsubsection{Comparison in terms of score selection} \textbf{With individual measures:} We compare the performance of models designated optimal by the advanced acceptance score and it's variants. 
The obtained results have been tabulated in Table \ref{tab:compVariants}, while we plot the corresponding radar plots in Fig. \ref{fig:MS_Inter_Radar}.
We find that $A_{r}^*(\Delta)$ (or $nA_{r}^*(\Delta)$ here) selects that model(/scores) which jointly satisfies all the design criterion. 
This is observed consistently for all the three datasets. 
On the other hand, the individual evaluation measures ($r,~\mathcal{R},~\Psi,~\&~C_d$) fail to do so. 
This is specifically noted in the Soli dataset. 
Although good gesture ranks and output scores were derived by our proposed $\mathcal{R}$. 
It was not sufficient to capture entanglement. 
This highlights the importance of multi-aspect evaluation in $A_{r}^*(\Delta)$. 
The SOTA acceptance score \cite{verma2024quantifying} also does not covers entanglement (see Fig. \ref{fig:MS_Inter_Radar} (a)) and trend deviation (see Fig. \ref{fig:MS_Inter_Radar} (a) $\&$ (b)). 
For the selected models, in Fig. \ref{fig:MS_QA}, we plot: (i) histogram comparing DGBQA scores and ground truth, and (ii) Gram matrix to visualize entanglement. 
The same highlights that in contrast to other models, the score/model selected by $A_{r}^*(\Delta)$ maintains the trade-off between lower entanglement (lighter the Gram matrix, the better it is) and high quality DGBQA scores (scores proportional to ground truth). 
It can be observed from Fig. \ref{fig:MS_QA} ((a), (b), and (f)), that the DGBQA scores of relevance and advanced acceptance score generate are more aligned with ground truth scores, in comparison to rank deviation. This proves the prominence of considering score quality during assessment. 

\textbf{With combined measures:} Combination of measures (Table \ref{tab:compVariants}, row 7-11) either selects the same model as $nA_{r}^*(\Delta)$, or the one which is not holistic (see Fig. \ref{fig:MS_Inter_Radar}).
It should be noted that these combined measures do not evaluate all the design criterion. 
As a result, selecting scores which is also selected by the $nA_{r}^*(\Delta)$ is because that score is optimal for all the criterion. This is not by the virtue of the evaluation measure. 
To verify this, reader can look into the case of $\bar{\psi}*\bar{C_d}$ vs $nA_{r}^*(\Delta)$ (Soli). Here, the selected score from MViT (0.5,1.0) was already optimal with respect to both entanglement and trend deviation, therefore its selection by $\bar{\psi}*\bar{C_d}$ was trivial. 
Another important comparison is between $\mathcal{R}*\bar{C_d}$ vs $nA_{r}^*(\Delta)$. 
With this comparison, we try to find insights upon importance of $\Psi$. Specifically, for the case of HandLogin and TinyRadar, we observe $\mathcal{R}*\bar{C_d}$ to be selecting that score which is suboptimal for trend deviation. Besides comparison, we observe that introducing a newer evaluation measure leads to making selection process influenced by the measure's design criterion. For instance, when we add $\bar{C_d}$ or $\bar{\Psi}$ in $A_{r}^*(\Delta)$, we notice the above mentioned change.

\textbf{Trade-off in TinyRadar dataset}: In the TinyRadar dataset, with $A_{r}^*(\Delta)$, although a lower entanglement and rank deviations were achieved. However, the relevance values also remain slightly lower than the other models. This is eminent for comparison with $\hat{r}$ and $\mathcal{R}$. Why is so? We identify two main reasons: (i) as described earlier, the entanglement process is suboptimal in this dataset, owing to which the values are not completely reliable. (ii) $A_{r}^*(\Delta)$ lays higher emphasis on entanglement. 

\begin{figure*}[!t]
    \centering
    \includegraphics[width=1.0\textwidth]{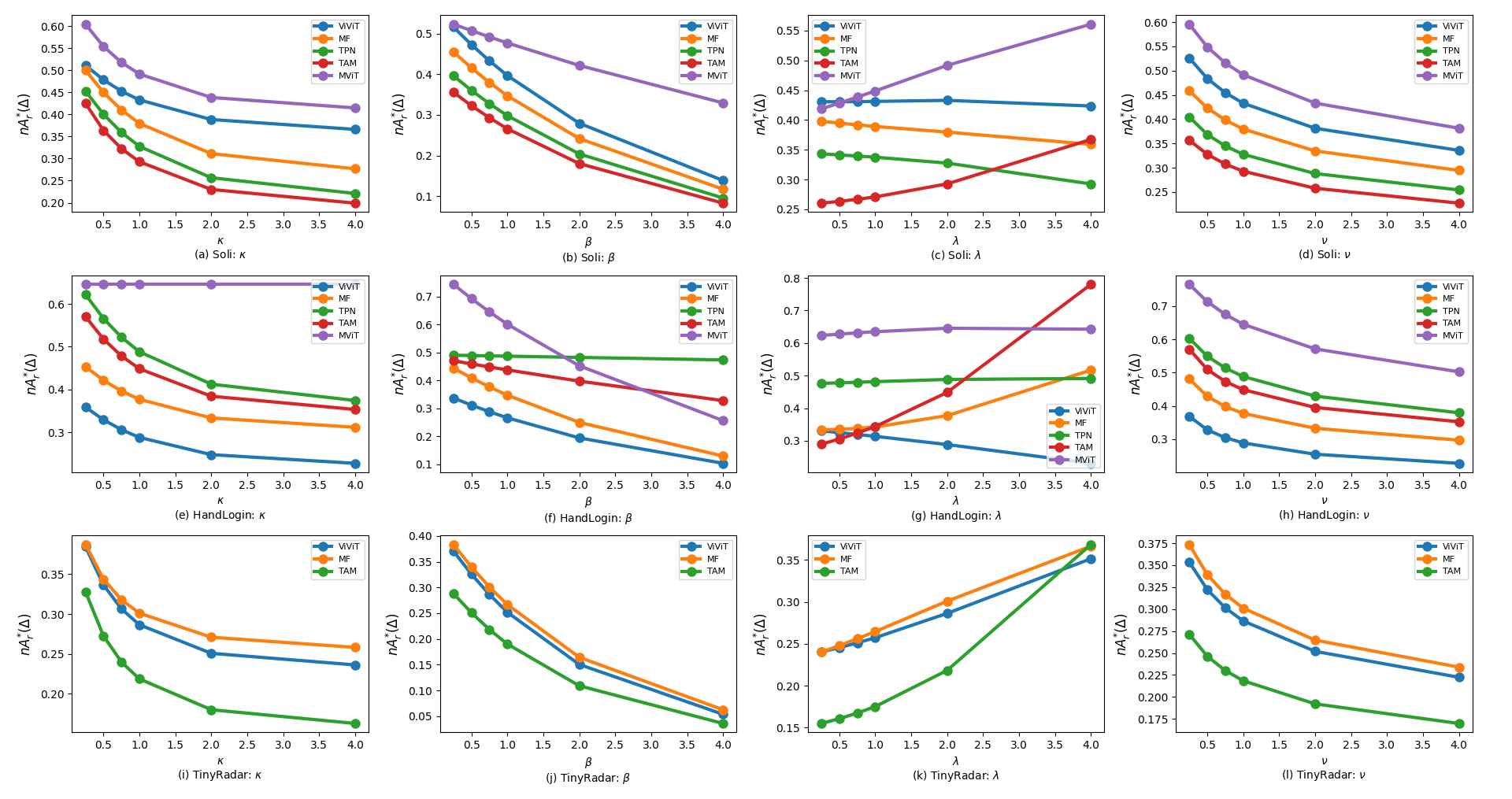}
    \caption{\textbf{Impact of varying scaling factors on $nA_{r}^*(\Delta)$}. We consider the optimal DGBQA for each of the dataset. Then, we plot $nA_{r}^*(\Delta)$ for different values of scaling factors. We vary a single factor at once. We observe that increasing or decreasing any of these factors results in significant changes in the $nA_{r}^*(\Delta)$. Thus, this validates: (i) advanced acceptance score is 
    sensitive to each of the individual parameters, and (ii) scaling factors can be utilized to select DGBQA scores as per user's preference.}
    \label{fig:HyperparamsBars}
\end{figure*}

\begin{figure*}[!t]
    \centering
    \includegraphics[width=1.0\textwidth]{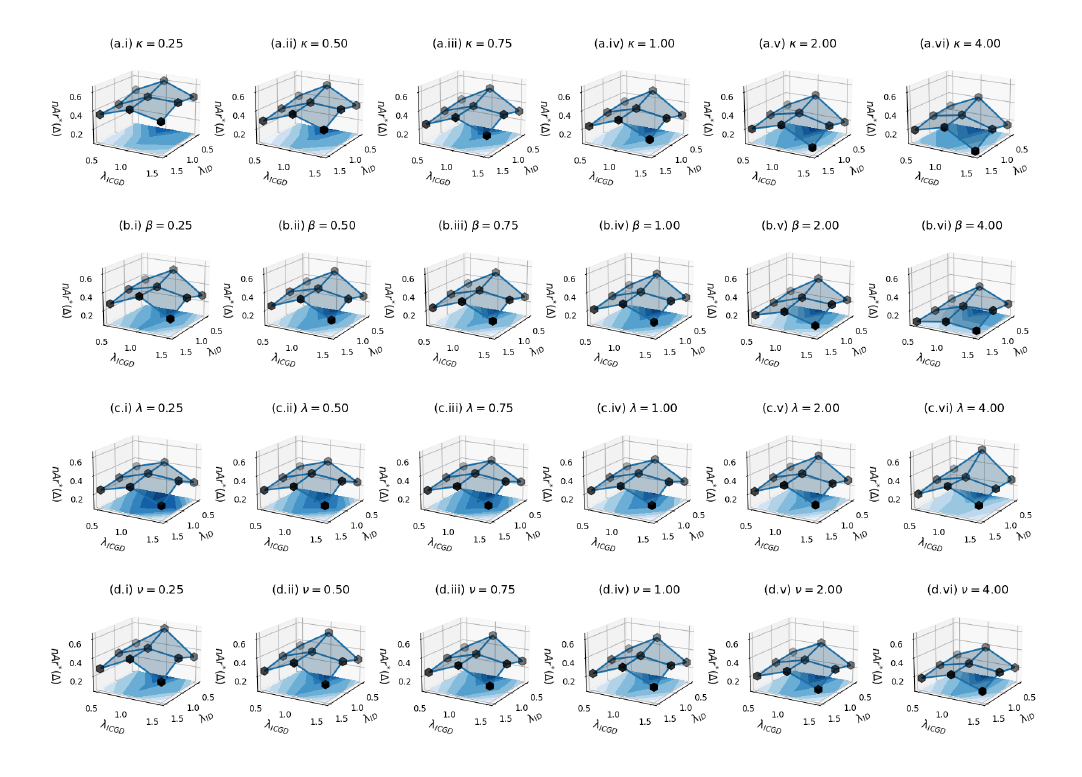}
    \caption{\textbf{Variation in surface plots of $nA_{r}^*(\Delta)$ vs $\lambda_{ID}~\&~\lambda_{ICGD}$ with changes in scaling factors.}
    Here, the surface represents $nA_{r}^*(\Delta)$ values of MViT models, trained on Soli using different $\lambda_{ID}~\&~\lambda_{ICGD}$. We study the changes in surface with varuation in scaling factors. The figure shows that increasing or decreasing any of these factors results in significant changes in the $nA_{r}^*(\Delta)$ (as also observed in Fig. \ref{fig:HyperparamsBars}).
    Further, when we compare across two surfaces across rows, we notice the entire surface getting raised or lowered. This suggests that individual evaluation measures can alter the score selection.}
    \label{fig:SurfacePlots}
\end{figure*}

\subsubsection{Comparison in terms of distribution of measures} \textbf{Are our design criterion measures conveying different information?} Theoretically yes, but we verify that the results also illustrate the same. To this end, we represent the distribution of measures using bar and whiskers plot (refer Fig. \ref{fig:MS_BarWhiskers}). With respect to individual parameters, each of them appear to measure different range of values. Also, they do not appear to follow similar distribution trends, while the median line is also different. This justifies our claims. An interesting case is that of $\Psi$ vs $\mathcal{R}$. We observed that a higher relevance leads to lower trend deviation (refer Table \ref{tab:compVariants}, row $2,3~\&~12$). 
Thus, we validate that $\Psi$ and $\mathcal{R}$ are different. We give four evidences: 
(i) as discussed in section \ref{sec:proposed}, they theoretically measure proportional yet different quantities. 
The former tries to seek if high ranks have higher DGBQA scores (and low for the lower ranks), while the latter quantifies the difference in score value progression between ground truth and DGBQA scores. 
(ii) From the bar and whisker's plot (refer Fig. \ref{fig:MS_BarWhiskers}, second and third bar), we can infer out that the $\Psi$ and $\mathcal{R}$ have overall distribution different. (iii) ViViT (0.5,2.5) trained on TinyRadar dataset gives a counter example. It attains a higher relevance value, but at the same time higher trend deviation as well. (iv) The score selection with $\mathcal{R}*\bar{C_d}$ did not ensure lower trend deviation in the case of HandLogin and TinyRadar dataset (refer Table \ref{tab:compVariants}, row $11$). Thus, this discussion establishes the importance of both the measures.

\textbf{Comparison with $A_{r}^*(\Delta)$}: We observe that the distribution of $A_{r}^*(\Delta)$ is more spread with respect to the individual measures. This highlights that combining all the individual measures results in a more discriminating measure ((refer Fig. \ref{fig:MS_BarWhiskers})). Next, we compare $A_{r}^*(\Delta)$ and $A_{r}^(\Delta)*\bar{C_d}$, as the latter results in almost the same selection as the former. Firstly, we observe some variability between their bars. This limited variation can be accredited to lower weightage (logarithm based formulation, refer eq. \ref{eq:Ar_star}) assigned to the trend deviation. Also, since each of the individual measures captures a different information, results from $A_{r}^*(\Delta)$ are generated considering more evaluation facets. This ensures higher reliability. 

\subsection{Comparison of the proposed with SOTA}
In this section we present comparison of our proposed measures with the one existing in SOTA. To this end, we consider the measure that can be adapted for our problem. Specifically, we take both standard measures: RMSE, cosine distance ($cos$), DCG, Kendall's rank inversion distance ($\tau$); and the recent ones: GRE \cite{lioma2017evaluation}, RPP \cite{diaz2022offline} for comparison. We have not considered the measures that: (i) utilize additional information not compatible with DGBQA scores \cite{sinha2023findability, ucar2024exploring, amigo2022ranking, tang2017investigating, raj2022measuring, rampisela2024can} and (ii) use priors which are not compatible with our problem \cite{maistro2021principled, de2021fairness}. For implementation details, please refer our code: (\href{https://github.com/AmanVerma2307/MeasureSuite}{https://github.com/AmanVerma2307/MeasureSuite}). The obtained results have been tabulated in Table \ref{tab:compSOTA}. While a pictorial representation of the same has been shown in Fig. \ref{fig:MS_Comp_Radar}.

\vspace{0.10cm}

\subsubsection{Comparison in terms of score selection} \textbf{$A_{r}^*(\Delta)$ vs SOTA}: We observe that $A_{r}^*(\Delta)$ selects the score which is jointly satisfies all the design criterion (refer Fig. \ref{fig:MS_Comp_Radar}). Apart from attaining high relevance and low rank deviation, it also obtains significantly lower entanglement. Existing measures does not portray this behavior. It is worthwhile noting that except the proposed 
$A_{r}^*(\Delta)$ neither of the measures, show optimality in score selection across all the three datasets. Reader can look into the case of RPP \cite{diaz2022offline} (refer Table \ref{tab:compSOTA}, row 12). It achieved good value in terms of relevance, but it falls short in covering a good trend deviation and entanglement (for HandLogin and TinyRadar dataset). Similarly, GRE \cite{lioma2017evaluation} and $A_{r}(\Delta)$ \cite{verma2024quantifying} focus majorly on rank deviation, while ignoring the other criterion, especially the trend deviation. Thus, we conclude that including all possible facets leads to more robust and exhaustive evaluation. 

\vspace{0.10cm}

\textbf{$\mathcal{R}$ vs SOTA}: The existing measures does not account for entanglement. Rather they assess the score quality. 
To this end, we compare our score quality measure: $\mathcal{R}$ (relevance) with them. 
For analysis we compare the selected models on the basis of relevance, trend deviation, and ranking order. 
It is evident from the Table \ref{tab:compSOTA} that $\mathcal{R}$ outperforms other measures across all the three datasets. 
For discounting based measures \cite{jarvelin2002cumulated, chapelle2009expected, gienapp2020impact}, we find that their obtained relevance values are suboptimal. 
This is because they do not account for lower low ranking scores. 
The rank based measures \cite{lioma2017evaluation, yilmaz2008estimating} fail to capture the trend deviation. This can be accredited to the fact that they emphasize on order of ranking, in contrast to quality of scores.
While the similarity based (RMSE,  $cos$, and $\tau$) results in lower trend deviation, but they do not ensure relevance and ranking order. However, as observed in Table \ref{tab:compVariants}, we find that these measures do not select the scores with lowest trend deviation (look into the results of Soli and TinyRadar). This validates that our proposed $\Psi$ is necessary to capture trend deviation. Thus, we establish the prominence of $\Psi$, $C_d$, and $\mathcal{R}$. When these are three are integrated with rank deviation, the resultant $A_{r}^*(\Delta)$ facilitates holistic evaluation.

\vspace{0.10cm}

\subsubsection{Correlation analysis} In Fig. \ref{fig:Correlation}, we present plots to study correlation between our proposed $nA_{r}^*(\Delta)$ and existing measures. We consider the best performing evaluation measures for this study.
As illustrated in the same, we find that the outputs of $nA_{r}^*(\Delta)$ are positively correlated with measures that are optimal at maximum value (DCG, cosine similarity, InfAp, RPP). While DCG is negatively correlated with global rank error (GRE). The slight deviation from linear fit is observed because, unlike existing measures, $nA_{r}^*(\Delta)$ also includes trend deviation and entanglement. Nevertheless, the $nA_{r}^*(\Delta)$ are higher for the scores, which other measures also claim to be higher. This suggest that our proposed advanced acceptance score is a reliable measure. 

\subsection{Ablation on scaling factors}
In this section we analyze the impact of changing scaling factors. We vary one of the factors ($\kappa$, $\beta$, $\lambda$, and $\nu$) at a time while keeping the others fixed. The choice of values are $\{0.25,0.50,0.75,1.00,2.00,4.00\}$. We perform this investigation in two fronts: (i) at inter-architecture level and (ii) wrt $\lambda_{ID}~\&~\lambda_{ICGD}$ at intra-architecture level. Firstly, we plot the $nA_{r}^*(\Delta)$ corresponding to optimal scores from different architectures for different values of scaling parameters (refer Fig. \ref{fig:HyperparamsBars}). 
Since we have assigned exponential weightage to entanglement (refer eq \ref{eq:Ar_star}), varying $\beta$ leads to highest decrement in $nA_{r}^*(\Delta)$ values. 
This is observed across all the datasets and architectures.
Hence, we conclude that the advanced acceptance score is highly sensitive to entanglement. 
As expected, we also find decrease in $nA_{r}^*(\Delta)$ with increasing $\kappa~\text{and}~\nu$. 
For higher value of $\lambda$ we find a dip in $nA_{r}^*(\Delta)$, this is because $A_{r}^*(\hat{e})$ also dominates with increase in $\lambda$ values. 
Thus, this plot (Fig. \ref{fig:HyperparamsBars}) highlights by changing the scaling factor, one can set preferences for the DGBQA scores. 
In particular, we provide these scaling factors for the users to select the score as per their desiderata. 
However, the values used by us in our experiments were empirically helpful in jointly satisfying all the design criterion. 
Next, we consider $\lambda_{ID}~\&~\lambda_{ICGD}$ values into account. To conduct this study, we consider MViT model trained on Soli for different $\lambda_{ID}~\&~\lambda_{ICGD}$ values. Then, we analyze the variation in their $nA_{r}^*(\Delta)$ values wrt scaling factors. We visually illustrate this in Fig. \ref{fig:SurfacePlots}. The inference drawn from Fig. \ref{fig:HyperparamsBars} are evident from here as well. 
Specifically, variation in $\beta$ significantly changes the $nA_{r}^*(\Delta)$ values. 
For all the scores, we find respective increment/decrement by changing $\kappa$, $\lambda$, and $\nu$. This is another evidence that our scaling parameters can be instrumental in setting preferences. Furthermore, this justifies the influence of individual measures for score selection. 

\section{Conclusion}
\label{sec:conclusion}
In this work, we proposed a holistic evaluation measure, the advanced acceptance score. It can be utilized as standalone measure for benchmarking gesture biometric characteristics quantification frameworks. 
Results illustrated that it can select scores which jointly satisfy the required design criterion. 
Specifically, the relevance measure, in contrast to other score quality metrics, rewards for low score of low ranking gestures. While the trend match distance, gave a novel method to compare score progressions. 
However, it consider only neighboring ranks. In future work, we will extend this to joint global trend measurement. 
Besides, we will also look into better fusion frameworks for the measures. The proposed measures are generic, and hence can be instrumental for capacity estimation in other biometric modalities. Furthermore, they can also be utilized in other problems involving graded retrieval and trend matching.

\bibliographystyle{IEEEtran}
\bibliography{main}{}

\begin{thebibliography}{10}
\providecommand{\url}[1]{#1}
\csname url@samestyle\endcsname
\providecommand{\newblock}{\relax}
\providecommand{\bibinfo}[2]{#2}
\providecommand{\BIBentrySTDinterwordspacing}{\spaceskip=0pt\relax}
\providecommand{\BIBentryALTinterwordstretchfactor}{4}
\providecommand{\BIBentryALTinterwordspacing}{\spaceskip=\fontdimen2\font plus
\BIBentryALTinterwordstretchfactor\fontdimen3\font minus \fontdimen4\font\relax}
\providecommand{\BIBforeignlanguage}[2]{{%
\expandafter\ifx\csname l@#1\endcsname\relax
\typeout{** WARNING: IEEEtran.bst: No hyphenation pattern has been}%
\typeout{** loaded for the language `#1'. Using the pattern for}%
\typeout{** the default language instead.}%
\else
\language=\csname l@#1\endcsname
\fi
#2}}
\providecommand{\BIBdecl}{\relax}
\BIBdecl

\bibitem{mendels2014user}
O.~Mendels, H.~Stern, and S.~Berma, ``{U}ser {I}dentification for {H}ome {E}ntertainment based on {F}ree-{A}ir {H}and {M}otion {S}ignatures,'' \emph{IEEE Transactions on Systems, Man, and Cybernetics: Systems}, vol.~44, no.~11, pp. 1461--1473, 2014.

\bibitem{kong2020continuous}
H.~Kong \emph{et~al.}, ``{C}ontinuous {A}uthentication through {F}inger {G}esture {I}nteraction for {S}mart {H}omes using {W}i{F}i,'' \emph{IEEE Transactions on Mobile Computing}, vol.~20, no.~11, pp. 3148--3162, 2020.

\bibitem{song2023understanding}
W.~Song, W.~Kang, and Y.~Zhang, ``{U}nderstanding {P}hysiological and {B}ehavioral {C}haracteristics {S}eparately for {H}igh-{P}erformance {V}ideo-based {H}and {G}esture {A}uthentication,'' \emph{IEEE Transactions on Instrumentation and Measurement}, vol.~72, pp. 1--13, 2023.

\bibitem{zhang2024robust}
Y.~Zhang, W.~Kang, and W.~Song, ``{R}obust and {A}ccurate {H}and {G}esture {A}uthentication with {C}ross-{M}odality {L}ocal-{G}lobal {B}ehavior {A}nalysis,'' \emph{IEEE Transactions on Information Forensics and Security}, 2024.

\bibitem{xie2023msba}
H.~Xie, W.~Song, and W.~Kang, ``{MSBA}-{N}et: {M}ulti-{S}cale {B}ehavior {A}nalysis {N}etwork for {R}andom {H}and {G}esture {A}uthentication,'' \emph{IEEE Transactions on Instrumentation and Measurement}, 2023.

\bibitem{verma2024quantifying}
A.~Verma \emph{et~al.}, ``{Q}uantifying {B}iometric {C}haracteristics of {H}and {G}estures {T}hrough {F}eature {S}pace {P}robing and {I}dentity-{L}evel {C}ross-{G}esture {D}isentanglement,'' in \emph{Proc. IEEE International Conference on Automatic Face and Gesture Recognition (FG)}, 2024, pp. 1--9.

\bibitem{terhorst2022limited}
P.~Terh{\"o}rst \emph{et~al.}, ``{O}n the (limited) {G}eneralization of {M}asterface {A}ttacks and its relation to the {C}apacity of {F}ace {R}epresentations,'' in \emph{Proc. IEEE International Joint Conference on Biometrics (IJCB)}, 2022, pp. 1--9.

\bibitem{boddeti2023biometric}
V.~N. Boddeti, G.~Sreekumar, and A.~Ross, ``{O}n the {B}iometric {C}apacity of {G}enerative {F}ace {M}odels,'' in \emph{Proc. IEEE International Joint Conference on Biometrics (IJCB)}, 2023.

\bibitem{ochoa2008relevance}
X.~Ochoa and E.~Duval, ``{R}elevance {R}anking {M}etrics for {L}earning {O}bjects,'' \emph{IEEE Transactions on Learning Technologies}, vol.~1, no.~1, pp. 34--48, 2008.

\bibitem{chapelle2009expected}
O.~Chapelle \emph{et~al.}, ``{E}xpected {R}eciprocal {R}ank for {G}raded {R}elevance,'' in \emph{Proc. ACM International Conference on Information and Knowledge Management}, 2009, pp. 621--630.

\bibitem{sakai2013summaries}
T.~Sakai and Z.~Dou, ``{S}ummaries, {R}anked {R}etrieval and {S}essions: {A} {U}nified {F}ramework for {I}nformation {A}ccess {E}valuation,'' in \emph{Proc. ACM SIGIR Conference on Research and Development in Information Retrieval}, 2013, pp. 473--482.

\bibitem{lioma2017evaluation}
C.~Lioma, J.~G. Simonsen, and B.~Larsen, ``{E}valuation {M}easures for {R}elevance and {C}redibility in {R}anked {L}ists,'' in \emph{Proc. ACM SIGIR Conference on Research and Development in Information Retrieval}, 2017, pp. 91--98.

\bibitem{gienapp2020estimating}
L.~Gienapp \emph{et~al.}, ``{E}stimating {T}opic {D}ifficulty using {N}ormalized {D}iscounted {C}umulated {G}ain,'' in \emph{Proc. ACM International Conference on Information and Knowledge Management}, 2020, pp. 2033--2036.

\bibitem{maistro2021principled}
M.~Maistro \emph{et~al.}, ``{P}rincipled {M}ulti-{A}spect {E}valuation {M}easures of {R}ankings,'' in \emph{Proc. ACM International Conference on Information and Knowledge Management}, 2021, pp. 1232--1242.

\bibitem{li2024evaluation}
X.~Li \emph{et~al.}, ``{O}n {E}valuation {M}etrics for {D}iversity-enhanced {R}ecommendations,'' in \emph{Proc. ACM International Conference on Information and Knowledge Management}, 2024, pp. 1286--1295.

\bibitem{jarvelin2002cumulated}
K.~J{\"a}rvelin and J.~Kek{\"a}l{\"a}inen, ``{C}umulated {G}ain-based {E}valuation of {IR} {T}echniques,'' \emph{ACM Transactions on Information Systems (TOIS)}, vol.~20, no.~4, pp. 422--446, 2002.

\bibitem{gienapp2020impact}
L.~Gienapp \emph{et~al.}, ``{T}he {I}mpact of {N}egative {R}elevance {J}udgments on n{DCG},'' in \emph{Proc. ACM International Conference on Information and Knowledge Management}, 2020, pp. 2037--2040.

\bibitem{sinha2023findability}
A.~Sinha, P.~R. Mall, and D.~Roy, ``{F}indability: {A} {N}ovel {M}easure of {I}nformation {A}ccessibility,'' in \emph{Proc. ACM International Conference on Information and Knowledge Management}, 2023, pp. 4289--4293.

\bibitem{valcarce2020assessing}
D.~Valcarce \emph{et~al.}, ``{A}ssessing {R}anking {M}etrics in {T}op-{N} {R}ecommendation,'' \emph{Information Retrieval Journal}, vol.~23, pp. 411--448, 2020.

\bibitem{moffat2008rank}
A.~Moffat and J.~Zobel, ``{R}ank-biased {P}recision for {M}easurement of {R}etrieval {E}ffectiveness,'' \emph{ACM Transactions on Information Systems (TOIS)}, vol.~27, no.~1, pp. 1--27, 2008.

\bibitem{amigo2022ranking}
E.~Amig{\'o}, S.~Mizzaro, and D.~Spina, ``{R}anking {I}nterruptus: {W}hen {T}runcated {R}ankings {A}re {B}etter and {H}ow to {M}easure {T}hat,'' in \emph{Proc. ACM SIGIR Conference on Research and Development in Information Retrieval}, 2022, pp. 588--598.

\bibitem{zunino2017revisiting}
A.~Zunino, J.~Cavazza, and V.~Murino, ``{R}evisiting {H}uman {A}ction {R}ecognition: {P}ersonalization vs. {G}eneralization,'' in \emph{Proc. International Conference on Image Analysis and Processing}, 2017, pp. 469--480.

\bibitem{de2021fairness}
T.~de~Freitas~Pereira and S.~Marcel, ``{F}airness in {B}iometrics: {A} {F}igure of {M}erit to {A}ssess {B}iometric {V}erification {S}ystems,'' \emph{IEEE Transactions on Biometrics, Behavior, and Identity Science}, vol.~4, no.~1, pp. 19--29, 2021.

\bibitem{dorsch2024fairness}
A.~D{\"o}rsch \emph{et~al.}, ``{F}airness {M}easures for {B}iometric {Q}uality {A}ssessment,'' \emph{arXiv preprint arXiv:2408.11392}, 2024.

\bibitem{van2024novel}
T.~V. Hamme \emph{et~al.}, ``{A} {N}ovel {E}valuation {F}ramework for {B}iometric {S}ecurity: {A}ssessing {G}uessing {D}ifficulty as a {M}etric,'' \emph{IEEE Transactions on Information Forensics and Security}, 2024.

\bibitem{garg2024gestformer}
M.~Garg \emph{et~al.}, ``{G}est{F}ormer: {M}ultiscale {W}avelet {P}ooling {T}ransformer {N}etwork for {D}ynamic {H}and {G}esture {R}ecognition,'' in \emph{Proc. IEEE/CVF Conference on Computer Vision and Pattern Recognition Workshops (CVPR-W)}, 2024, pp. 2473--2483.

\bibitem{kopuklu2020online}
O.~K{\"o}p{\"u}kl{\"u} \emph{et~al.}, ``{O}nline {D}ynamic {H}and {G}esture {R}ecognition including {E}fficiency {A}nalysis,'' \emph{IEEE Transactions on Biometrics, Behavior, and Identity Science}, vol.~2, no.~2, pp. 85--97, 2020.

\bibitem{yoon2020speech}
Y.~Yoon \emph{et~al.}, ``{S}peech {G}esture {G}eneration from the {T}rimodal {C}ontext of {T}ext, {A}udio, and {S}peaker {I}dentity,'' \emph{ACM Transactions on Graphics (TOG)}, vol.~39, no.~6, pp. 1--16, 2020.

\bibitem{voas2023best}
J.~Voas \emph{et~al.}, ``{W}hat is the {B}est {A}utomated {M}etric for {T}ext to {M}otion {G}eneration?'' in \emph{Proc. SIGGRAPH Asia}, 2023, pp. 1--11.

\bibitem{yilmaz2008estimating}
E.~Yilmaz and J.~A. Aslam, ``{E}stimating {A}verage {P}recision {w}hen {J}udgments are {I}ncomplete,'' \emph{Knowledge and Information Systems}, vol.~16, no.~2, pp. 173--211, 2008.

\bibitem{tang2017investigating}
Z.~Tang and G.~H. Yang, ``{I}nvestigating per {T}opic {U}pper {B}ound for {S}ession {S}earch {E}valuation,'' in \emph{Proc. ACM SIGIR International Conference on Theory of Information Retrieval}, 2017, pp. 185--192.

\bibitem{jeunen2024normalised}
O.~Jeunen, I.~Potapov, and A.~Ustimenko, ``{O}n (normalised) {D}iscounted {C}umulative {G}ain as an {O}ff-{P}olicy {E}valuation {M}etric for {T}op-{N} {R}ecommendation,'' in \emph{Proc. ACM SIGKDD Conference on Knowledge Discovery and Data Mining (KDD)}, 2024, pp. 1222--1233.

\bibitem{amigo2018axiomatic}
E.~Amig{\'o}, D.~Spina, and J.~C. de~Albornoz, ``{A}n {A}xiomatic {A}nalysis of {D}iversity {E}valuation {M}etrics: {I}ntroducing the {R}ank-{B}iased {U}tility {M}etric,'' in \emph{Proc. ACM SIGIR Conference on Research and Development in Information Retrieval}, 2018, pp. 625--634.

\bibitem{diaz2022offline}
F.~Diaz and A.~Ferraro, ``{O}ffline {R}etrieval {E}valuation without {W}valuation {M}etrics,'' in \emph{Proc. ACM SIGIR Conference on Research and Development in Information Retrieval}, 2022, pp. 599--609.

\bibitem{jaswal2021range}
G.~Jaswal, S.~Srirangarajan, and S.~{Dutta~Roy}, ``{R}ange-{D}oppler {H}and {G}esture {R}ecognition using {D}eep {R}esidual-3{D} {T}ransformer {N}etwork,'' in \emph{Pattern Recognition. ICPR International Workshops and Challenges}, 2021, pp. 311--315.

\bibitem{arnab2021vivit}
A.~Arnab \emph{et~al.}, ``Vi{V}i{T}: {A} {V}ideo {V}ision {T}ransformer,'' in \emph{Proc. International Conference on Computer Vision (ICCV)}, 2021, pp. 6836--6846.

\bibitem{patrick2021keeping}
M.~Patrick \emph{et~al.}, ``{K}eeping {Y}our {E}ye on the {B}all: {T}rajectory {A}ttention in {V}ideo {T}ransformers,'' \emph{Proc. Advances in Neural Information Processing Systems (NeurIPS)}, pp. 12\,493--12\,506, 2021.

\bibitem{fan2021multiscale}
H.~Fan \emph{et~al.}, ``{M}ultiscale {V}ision {T}ransformers,'' in \emph{Proc. IEEE/CVF Conference on Computer Vision and Pattern Recognition (CVPR)}, 2021, pp. 6824--6835.

\bibitem{yang2020temporal}
C.~Yang \emph{et~al.}, ``{T}emporal {P}yramid {N}etwork for {A}ction {R}ecognition,'' in \emph{Proc. IEEE/CVF Conference on Computer Vision and Pattern Recognition (CVPR)}, 2020, pp. 591--600.

\bibitem{liu2021temporal}
Z.~Liu \emph{et~al.}, ``{T}emporal {A}daptive {M}odule for {V}ideo {R}ecognition,'' in \emph{Proc. International Conference on Computer Vision (ICCV)}, 2021, pp. 13\,688--13\,698.

\bibitem{wang2016interacting}
S.~Wang \emph{et~al.}, ``{I}nteracting with {S}oli: {E}xploring {F}ine-{G}rained {D}ynamic {G}esture {R}ecognition in the {R}adio-{F}requency {S}pectrum,'' in \emph{Proc. Symposium on User Interface Software and Technology}, 2016, pp. 851--860.

\bibitem{wu2015leveraging}
J.~Wu \emph{et~al.}, ``{L}everaging {S}hape and {D}epth in {U}ser {A}uthentication from {I}n-{A}ir {H}and {G}estures,'' in \emph{Proc. IEEE International Conference on Image Processing (ICIP)}, 2015, pp. 3195--3199.

\bibitem{sheng2022progressive}
X.~Sheng \emph{et~al.}, ``{A} {P}rogressive {D}ifference {M}ethod for {C}apturing {V}isual {T}empos on {A}ction {R}ecognition,'' \emph{IEEE Transactions on Circuits and Systems for Video Technology}, vol.~33, no.~3, pp. 977--987, 2022.

\bibitem{scherer2021tinyradarnn}
M.~Scherer \emph{et~al.}, ``{TinyRadarNN}: {C}ombining {S}patial and {T}emporal {C}onvolutional {N}eural {N}etworks for {E}mbedded {G}esture {R}ecognition with {S}hort {R}ange {R}adars,'' \emph{IEEE Internet Things J.}, vol.~8, no.~13, pp. 10\,336--10\,346, 2021.

\bibitem{ucar2024exploring}
T.~Ucar, C.~Malherbe, and F.~Gonzalez, ``{E}xploring {L}og-{L}ikelihood {S}cores for {R}anking {A}ntibody {S}equence {D}esigns,'' in \emph{Proc. NeurIPS Workshop on AI for New Drug Modalities}, 2024.

\bibitem{raj2022measuring}
A.~Raj and M.~D. Ekstrand, ``{M}easuring {F}airness in {R}anked {R}esults: {A}n {A}nalytical and {E}mpirical {C}omparison,'' in \emph{Proc. ACM SIGIR Conference on Research and Development in Information Retrieval}, 2022, pp. 726--736.

\bibitem{rampisela2024can}
T.~V. Rampisela \emph{et~al.}, ``{C}an {W}e {T}rust {R}ecommender {S}ystem {F}airness {E}valuation? {T}he {R}ole of {F}airness and {R}elevance,'' in \emph{Proc. ACM SIGIR Conference on Research and Development in Information Retrieval}, 2024, pp. 271--281.

\end{thebibliography}
\end{document}